\documentclass{article}

\usepackage{microtype}
\usepackage{graphicx}
\usepackage{subcaption}
\usepackage{booktabs} 

\usepackage{hyperref}

\usepackage[preprint]{icml2026}

\usepackage{amsmath}
\usepackage{amssymb}
\usepackage{mathtools}
\usepackage{amsthm}

\usepackage{graphicx}
\usepackage{amsmath}
\usepackage{amsfonts}
\usepackage{svg}
\usepackage[utf8]{inputenc}
\usepackage[T1]{fontenc}
\usepackage{listings}
\usepackage{xcolor}
\usepackage{amssymb} 
\usepackage{mathtools}
\usepackage{algpseudocode}

\usepackage{latexsym}
\usepackage{adjustbox}
\usepackage{booktabs}
\usepackage{multirow}
\usepackage{xcolor}
\usepackage{colortbl}
\usepackage{natbib}
\usepackage{subcaption}

\usepackage{tikz}
\usetikzlibrary{arrows.meta, shapes.geometric, positioning, calc, decorations.pathmorphing, patterns, backgrounds}

\usepackage{amsmath, amssymb}
\newcommand{\Romannum}[1]{%
  \uppercase\expandafter{\romannumeral #1}%
}
\newcommand{\romannum}[1]{%
  \romannumeral #1%
}

\usepackage{tcolorbox}
\usepackage{tabularx}
\usepackage{xcolor}

\usepackage{fontawesome5} 

\makeatother

\usepackage[capitalize,noabbrev]{cleveref}

\theoremstyle{plain}

\theoremstyle{definition}

\theoremstyle{remark}

\usepackage[textsize=tiny]{todonotes}

\icmltitlerunning{Budget-Constrained Agentic Large Language Models}

\begin{document}

\twocolumn[
  \icmltitle{Budget-Constrained Agentic Large Language Models: \\ Intention-Based Planning for Costly Tool Use}



  \icmlsetsymbol{equal}{*}

  \begin{icmlauthorlist}
    \icmlauthor{Hanbing Liu}{ruc}
    \icmlauthor{Chunhao Tian}{ruc}
    \icmlauthor{Nan An}{ruc}
    \icmlauthor{Ziyuan Wang}{ruc}
    \icmlauthor{Pinyan Lu}{sufe}
    \icmlauthor{Changyuan Yu}{baidu}
    \icmlauthor{Qi Qi}{ruc}
\end{icmlauthorlist}

  \icmlaffiliation{ruc}{Gaoling School of Artificial Intelligence, Renmin University of China}
  \icmlaffiliation{sufe}{Shanghai University of Finance and Economics}
  \icmlaffiliation{baidu}{Baidu Inc.}

  \icmlcorrespondingauthor{Qi Qi}{qi.qi@ruc.edu.cn}

  \icmlkeywords{Machine Learning, ICML}
  \vskip 0.3in
]



\printAffiliationsAndNotice{}  

\begin{abstract}
  We study budget-constrained tool-augmented agents, where a large language model must solve multi-step tasks by invoking external tools under a strict monetary budget.
  We formalize this setting as sequential decision making in context space with priced and stochastic tool executions, making direct planning intractable due to massive state–action spaces, high variance of outcomes and prohibitive exploration cost.

To address these challenges, we propose INTENT, an inference-time planning framework that leverages an intention-aware hierarchical world model to anticipate future tool usage, risk-calibrated cost, and guide decisions online. Across cost-augmented StableToolBench, INTENT strictly enforces hard budget feasibility while substantially improving task success over baselines, and remains robust under dynamic market shifts such as tool price changes and varying budgets.
\end{abstract}

\section{Introduction}\label{sec:intro}
Large language models are rapidly evolving into agentic systems that can autonomously decompose complex tasks, interact with external tools, and execute multi-step plans. Recent systems demonstrate strong capabilities in deep research \cite{openai2025deepresearch, team2025tongyi}, software engineering \cite{yang2024swe}, and web automation \cite{google2026}, where reasoning and tool use are tightly interleaved.

Tools constitute the primary interface through which agents interact with the external world and incur real-world impact. With the emergence of standardized protocols such as MCP \cite{anthropic_mcp} and large-scale tool marketplaces \cite{rapid2014,mcpmarket2025,mcpso2025}, agents now have access to thousands of heterogeneous APIs. While this dramatically expands the action space, it also introduces a critical but underexplored dimension: \textsc{economic cost}. Unlike token generation, whose marginal cost continues to decline drastically \cite{a16z2024,epoch2025}, many tools expose scarce and monetized resources, such as real-time financial market feeds, high-resolution satellite imagery or irrevocable blockchain state changes.

\begin{figure}[t]
\centering
\includegraphics[width=\linewidth]{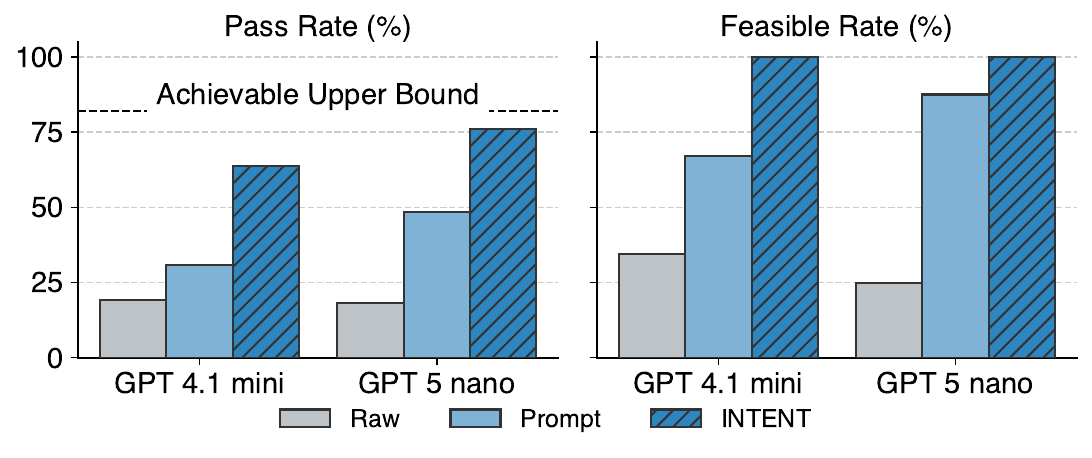}
\caption{Budget awareness of agentic language models on tool cost-augmented StableToolBench.
Standalone agents frequently violate hard budget constraints, and prompt-based cost feedback remains insufficient to guarantee budget feasibility or approach the achievable performance upper bound. Our lightweight online planning framework INTENT helps bridge this gap.}
\label{fig:banner}
\end{figure}
As agentic systems mature, the central question is no longer whether they can solve complex tasks, but whether we can delegate economically consequential decisions to them. A fundamental question arises: 
\textbf{Can we trust agentic models to make cost-sensitive tool-use decisions on our behalf?}

To probe this question, we instantiate a budgeted tool-use setting on top of the widely used StableToolBench \cite{Qin2023ToolLLMFL,Guo2024StableToolBenchTS}, where each task comes with a hard budget constraint and diverse per-call tool prices. As shown in Figure~\ref{fig:banner}, our findings reveal a significant gap: even when explicitly provided with budget feedback after each tool call (\textsc{Prompt}), strong models frequently exceed the budget due to repetitive retries and unproductive exploration. More advanced reasoning models exhibit better compliance, but only by becoming overly conservative, leaving a large performance gap to the achievable upper bound.

{
Such budgeted tool-use scenarios are not rare in practice. 
For instance, a financial analyst may rely on agentic workflow to monitor hundreds of stocks on a daily basis, and, to control operational expenditure, impose a strict per-ticker spending limit when gathering evidence about fundamentals and public sentiment. 
For each ticker, the agent then faces a spectrum of priced tool set: near-free but noisy web search may require repeated retries, official filings databases are moderately priced yet brittle to query formulations, and premium data APIs are accurate but can consume most of the budget in one shot. 
In such cases, myopic choices like overspending early on expensive tools or looping on unreliable cheap calls can render the remaining subtasks infeasible even when each individual step appears reasonable. 
Beyond finance, similar cost-capped tool-use patterns arise in many real-world pipelines (e.g., legal due diligence and cloud incidence diagnosis), and become particularly pronounced when agents are deployed at scale and invoked repeatedly.
}

This setting exposes a non-trivial challenge. Agents must make sequential tool-use decisions under {high stochasticity}, where actions may incur real economic cost, tool outcomes are {uncertain}, and neither free interaction nor retraining is available at inference time. Moreover, the tool market itself is \textit{dynamic}: available tools and their prices {may change  with time}, and new tools may appear without prior experience.

Motivated by these observations and {examples}, in this work, we study a concrete instantiation of cost-sensitive tool use that captures these challenges while remaining amenable to principled analysis.
Each task consists of a user query and a task-specific tool market, where finite available tools have heterogeneous \textit{per-call }prices. The agent must decide \textit{which} tools to invoke, in \textit{what order}, and \textit{when} to terminate, so as to solve the task under a hard budget constraint.

At first glance, this problem appears amenable to several natural solution strategies. One may attempt to cast it as an online knapsack \cite{Buchbinder2009TheDO,Agrawal2009ADN} or linear programming variant, allocating the budget across tools based on their shadow prices. However, such formulations assume independent, additive utilities and fail to capture the strong sequential dependencies between tool calls, where the value of an action is primarily determined by the information it enables for subsequent decisions.

{
Alternatively, one might consider endowing agents with budget awareness through reinforcement learning in the post-training stage \cite{Schulman2017ProximalPO, Shao2024DeepSeekMathPT}.
However, this paradigm is fundamentally misaligned with our setting.
During training, an agent can only bind tool knowledge attained from historical interaction to static textual descriptions, while real-time factors that substantially affect \textit{effective value} of a tool, such as shifts in service quality, coverage, or response reliability,
are inherently unobservable and only revealed at inference time. Tracking such dynamics would therefore require frequent retraining, which is already prohibitively expensive for large agentic models, let alone the additional monetary cost incurred by on-policy data collection through real tool invocations.}

{Besides,} classical online planning methods such as Monte Carlo Tree Search \cite{Silver2010MonteCarloPI, Lee2018MonteCarloTS} offer a principled way to reason about long-horizon decisions under uncertainty. However, they typically rely on free environment interaction and extensive branching, resulting in prohibitive latency for agentic settings with {extremely} large action spaces and expensive tool calls.

Taken together, these limitations point to a narrow but crucial design space: a solution must operate purely at inference time, reason about future costs under stochastic tool outcomes, and remain lightweight enough to guide a strong pretrained agent without exhaustive search.

Thus, in this work, we propose INTENT, a lightweight inference-time planning framework for budget-aware tool use. INTENT leverages a learned language world model to simulate tool outcomes and performs calibrated Monte Carlo lookahead to estimate future costs. Crucially, INTENT introduces an intention-based decomposition that separates whether a tool call satisfies the agent’s semantic intention from the concrete content of the tool output, enabling accurate cost estimation in highly stochastic environments.

Our contributions are summarized as follows:
\begin{itemize}
    \item We formalize budget-constrained tool use as a sequential decision problem in agentic language models.
    \item We propose INTENT, an intention-based planning algorithm that enables budget-aware decision making without retraining or environment interaction.
    \item We demonstrate substantial performance improvements on StableToolBench across diverse budgets and market settings, approaching the empirical upper bound with mild overhead.
\end{itemize}

\section{Model}\label{sec:model}
In this section, We formalize budget-constrained agentic tool use as a sequential decision making over a growing textual history. Each task instance specifies a user query, a hard budget constraint, and a snapshot of a dynamic tool marketplace with per-call costs. The agent follows a interleaved thinking loop that alternates between reasoning, tool calls, and stochastic observations returned by external tools, until it terminates with a final answer. This abstraction captures two core challenges of real-world agentic systems: an unbounded action space induced by free-form arguments, and stochastic transitions arising from tool execution.

\subsection{Contextual State}


We represent the agent’s interaction with the environment as sequential decision making over a growing textual context.

\textbf{Context Space.}
Let $\mathcal{V}$ denote the discrete vocabulary of tokens, and let $\mathcal{V}^*$ denote the set of all finite token sequences.
We define the concatenation of two sequences $x_1, x_2 \in \mathcal{V}^*$ as $[x_1, x_2]$, and extend this notation naturally to multiple sequences $[x_1, x_2, \dots, x_n]$.
The agent’s state is represented implicitly by its entire interaction history, which grows monotonically through concatenation.

\textbf{Serialization Convention.}
Throughout this paper, abstract objects that the agent interacts with, such as a tool specification $T$ and execution feedback $o$, are assumed to admit a canonical textual serialization.
For simplicity, we identify each object with its serialized token sequence, and use the same symbol to denote both, 
whenever no ambiguity arises.
This convention allows us to embed heterogeneous concepts uniformly into the same context space.

\textbf{Language Model.}
The agent is powered by a large language model parameterized by $\theta$.
We view the LLM as a probabilistic policy operating over the context space.
Given a context sequence $h \in \mathcal{V}^*$, the probability of generating a continuation $x \in \mathcal{V}^*$ is denoted by
\(
P_\theta(x \mid h).
\)

This formulation enables us to treat reasoning traces, tool calls, and final answers uniformly as token sequences generated by a single policy over the contextual state.

\subsection{Dynamic Tool Marketplace}

We next formalize the environment in which the agent operates, stressing the dynamic availability and pricing of tools.

\textbf{Market Snapshot.}
We assume a universe of all potential tools $\mathcal{T}$.
Upon the arrival of each user query $q$, the agent is presented with a market snapshot, denoted by $\mathcal{M}$.
This snapshot specifies the subset of tools that are accessible at that moment (determined by retrieval mechanisms, user permissions, or provider status), together with their current \textit{per-call} costs.
Formally,
\(
\mathcal{M} = \{(T^{(j)}, c^{(j)})\}_{j=1}^m,
\)
where each $T^{(j)} \in \mathcal{T}$ encodes the tool’s specification (e.g., description, input schema, and usage examples), and $c^{(j)} \in \mathbb{R}_{\ge 0}$ denotes the financial cost incurred by a single invocation of the tool.

\textbf{Task Instance.}
A task instance is defined as a tuple $\mathcal{I} = (q, B, \mathcal{M})$, sampled from a task distribution $\mathcal{D}$.
Here, $q$ denotes the user’s natural language query, $B \in \mathbb{R}_{>0}$ is a hard budget constraint, and $\mathcal{M}$ specifies the tool market faced by the agent for this particular request.

\subsection{Budget-Constrained Agent}
Given a task instance $\mathcal{I} = (q, B, \mathcal{M})$, we model the agent’s problem-solving process as a ReAct-style~\cite{Yao2022ReActSR} sequential decision-making procedure over a growing textual history.
The interaction unfolds over discrete time steps $t = 1, 2, \dots$, during which the agent alternates between internal reasoning, external tool use, and observation.

\textbf{History Initialization.}
Rather than maintaining an abstract state representation, we treat the agent’s state as its full interaction history.
The initial history $h_0$ consists of the system prompt, the user query, the budget constraint, and the market snapshot:
\(
h_0 = [\texttt{System}, q, B, \mathcal{M}].
\)

\textbf{Reasoning and Action.}
At each step $t$, the agent first generates a reasoning trace $r_t$ to plan or reflect, conditioned on the current history,
\(
r_t \sim \pi_{\theta}(\cdot \mid h_t).
\)
Based on both the history and the reasoning trace, the agent then selects a structured action
\(
a_t \sim \pi_{\theta}(\cdot \mid [h_t, r_t]).
\)
The action space $\mathcal{A}$ consists of two types of operations:

(\romannum{1}) \textsc{Tool Call}. $a_t = (\textsc{Call}, T_t, u_t)$, where $T_t \in \{{T^{(j)}}\}_{j=1}^m$ is a tool available in the current market $\mathcal{M}$ and $u_t$ denotes its arguments. (\romannum{2}) \textsc{Termination}. $a_t = (\textsc{Answer}, y)$, where $y$ is the final response to the user.

\textbf{State Transition.}
If the agent chooses to terminate, the episode ends.
If the agent chooses to call a tool, the transition is governed by an external environment $\mathcal{E}$, which acts as a chance node.
The tool execution produces a \textit{stochastic} observation
\(
o_t \sim \mathcal{E}(\cdot \mid T_t, u_t),
\)
and incurs a cost $\textsc{Cost}(a_t) = c^{(j)}$, where $T_t = T^{(j)}$.
The history is then deterministically updated by appending the interaction block:
\(
h_{t+1} = [h_t, r_t, a_t, o_t].
\)

\subsection{Task Formulation}

A trajectory $\tau$ is defined as the sequence of interaction components generated until termination at step $K$,
\(
\tau = [h_0, r_1, a_1, o_1, \dots, a_K],
\)
where $a_K$ is a \textsc{Answer} action.

\textbf{Reward.}
The agent’s objective is to produce a high-quality answer while strictly satisfying the budget constraint.
Let $J(a_K, q) \in [0, 1]$ denote an external evaluation metric (e.g., a human evaluator or a judge model) that assesses whether the user query $q$ has been successfully resolved.
We define the reward function as
\[
R(\tau) = J(a_K, q) \cdot \mathbb{I}\!\left( \sum_{t=1}^{K} \textsc{Cost}(a_t) \le B \right),
\]
where $\mathbb{I}(\cdot)$ is an indicator function enforcing a hard budget constraint.
Trajectories that exceed the budget receive zero reward regardless of answer quality.

\textbf{Optimization Objective.}
Our goal is to find a policy $\pi_\theta$ that maximizes the expected reward over the task distribution:
\[
\max_{\pi} \; \mathbb{E}_{\mathcal{I} \sim \mathcal{D}}
\left[ \mathbb{E}_{\tau \sim \pi, \mathcal{E}} \big[ R(\tau) \big] \right].
\]
Directly optimizing this objective in post-training stage is impractical.
The action space induced by free-form tool arguments is extremely large, making exploration and credit assignment prohibitively expensive.
More importantly, the agent operates in a non-stationary tool market: available tools and their per-call costs vary across task instances, and new tools may appear without prior training data.
As a result, a policy optimized offline cannot reliably adapt to the market configuration faced at inference time.

These challenges motivate an online and lightweight inference-time planning approach.
Instead of modifying the agent’s parameters, we leverage a learned world model to approximate environment transitions and dynamically guide tool selection under budget constraints.


\section{Methodology}\label{sec:method}
\begin{figure*}[t]
\centering
\begin{subfigure}[t]{0.25\textwidth}
    \centering
    \includegraphics[width=\linewidth]{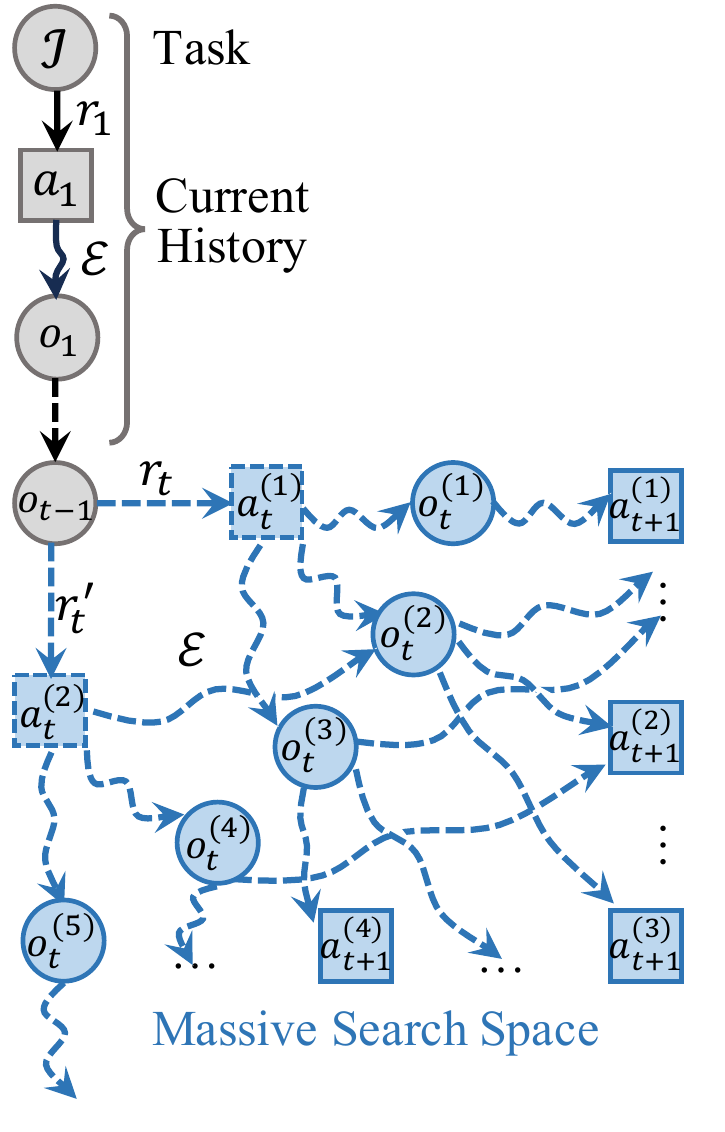}
    \caption{Monte Carlo Tree Search}
\end{subfigure}
\begin{subfigure}[t]{0.32\textwidth}
    \centering
    \includegraphics[width=\linewidth]{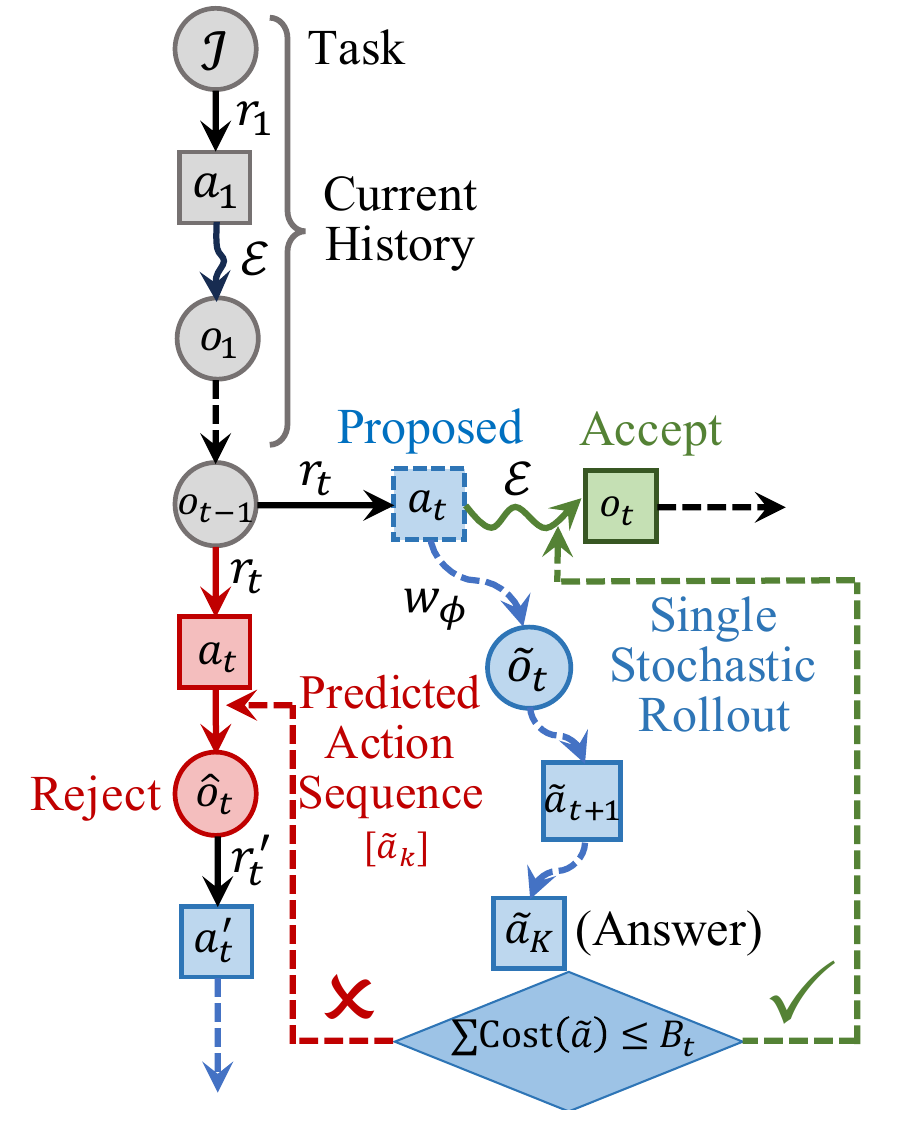}
    \caption{Monte Carlo Oracle (MCO)}
\end{subfigure}
\begin{subfigure}[t]{0.32\textwidth}
    \centering
    \includegraphics[width=\linewidth]{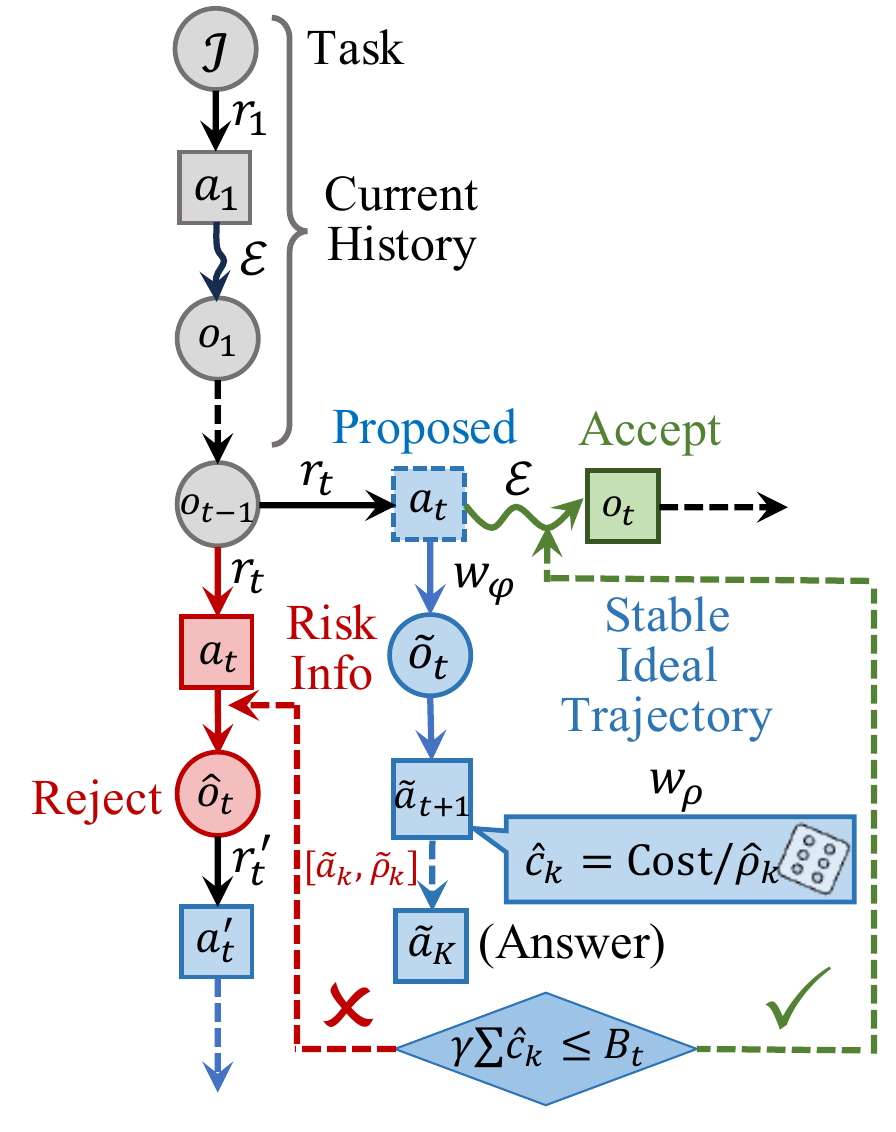}
    \caption{Intention-based Oracle (INTENT)}
\end{subfigure}
\caption{Inference-time planning paradigms for budget-aware agentic tool use. (a) MCTS explores a large stochastic search tree with prohibitive cost. (b) MCO enforces budgets via a single stochastic rollout using a language world model. (c) INTENT extracts the agent’s latent plan through ideal trajectory simulation and applies intention-aware, risk-adjusted cost estimation for stable budget control.}
\label{fig:framework}
\end{figure*}

We now introduce an inference-time planning framework for budget-constrained tool use.
Our design goal is to minimally intervene on a strong pretrained agent, while dynamically enforcing hard budget constraints under a non-stationary tool market.
To this end, 
we perform lightweight lookahead simulations using a learned world model to anticipate future tool usage and guide decision making online. This section focuses on the conceptual design of the method. Figure~\ref{fig:framework} outlines and contrasts candidate frameworks, while implementation details, including training and algorithmic pseudocode, are provided in the Appendix~\ref{app:algorithm}.



\subsection{Language World Model}
First, we train a world model $\mathcal{W}_\phi$, parameterized by $\phi$, to predict the outcome of tool executions. While LLMs may not perfectly simulate the factual accuracy of external tools (e.g., a specific stock price), they excel at predicting the \textit{format} and \textit{structure} of outputs, which is sufficient to elicit the agent's latent planning logic.
Given tool call $T_t \in \mathcal{M}$, and arguments $u_t$, the model predicts the observation:
$$
\tilde{o}_t \sim \mathcal{W}_\phi(\cdot \mid [T_t, u_t]).
$$
Importantly, we do not employ the language world model (LWM) for exhaustive tree search.
Classical planning algorithms such as Monte Carlo Tree Search \cite{hao2023reasoning,Zhou2023LanguageAT} rely on repeated branching and state reuse, which are infeasible in our setting due to the unbounded action space induced by free-form tool call arguments, and the high inference latency of agents.
Instead, we restrict the usage of world model to single-trajectory lookahead simulation, which is sufficient for enforcing budget constraints while remaining computationally lightweight.

\subsection{Monte Carlo Oracle}
Our direct feasible application of the LWM is the \textbf{Monte Carlo Oracle (MCO)}. The core philosophy is \textit{minimal intervention}: we assume the agent's policy $\pi_\theta$ is inherently capable of solving the task, and the planning algorithm's role is strictly limited to enforcing budget constraints.

\textbf{Mechanism.} 
When the agent proposes an immediate action $a_t$ under current context $[h_t, r_t]$ and remaining budget $B_t$, MCO performs a single \textit{Lookahead Rollout}. Starting from the current state $h_t$, we alternate between the world model $\mathcal{W}_\phi$ and the agent policy $\pi_\theta$ to generate a simulated future trajectory $
\tilde{\tau} = (a_t, \tilde{o}_t, \tilde{r}_{t+1}, \tilde{a}_{t+1}, \tilde{o}_{t+1}, \dots, \tilde{a}_K),
$
where $\tilde{a}_K$ is a terminal \textsc{Answer} action. 
We then evaluate the total projected cost $C(\tilde{\tau}) = \sum_{\tilde{a} \in \tilde{\tau}} \textsc{Cost}(\tilde{a})$.

\textbf{Decision and Feedback.}
We rely on the \textit{Capability Assumption}: if the agent decides to terminate at step $K$, we assume the gathered information is sufficient. The decision logic is purely budget-based: (\romannum{1}) \textbf{Accept.} If $C(\tilde{\tau}) \le B_{t}$, the action $a_t$ is allowed to be executed in the real environment; (\romannum{2}) \textbf{Reject.} If $C(\tilde{\tau}) > B_{t}$, the oracle intercepts $a_t$ and prevents its execution. Crucially, to guide the agent's re-planning, we construct a feedback observation $\hat{o}_t = [a_t, \tilde{a}_{t+1}, \dots, \tilde{a}_K ]$ containing the sequence of simulated actions that led to the budget violation. This feedback exposes the future failure to the agent, prompting it to generate a new reasoning trace $r'_t$ and a more informed action $a'_t$ via $\pi_{\theta}(\cdot \mid [h_t, r_t,a_t,\hat{o}_t]).$

\textbf{Limitation.} MCO relies on a single sample estimate. Due to the \textit{stochastic} nature of tools (e.g., a search engine might return irrelevant results, triggering a costly retry loop), the variance of $\textsc{Cost}(\tilde{\tau})$ is high. A single lucky simulation may underestimate the true expected cost, leading to budget overruns in deployment.
\subsection{Intention-Based Oracle}\label{subsec:intent}

To mitigate the high variance of single-sample estimation in MCO, we propose \textbf{INTENT}, which is motivated by the observation that an agent's decision to alter its high-level plan is driven less by the specific tool call outcome and more by whether it \textit{satisfies the intention} encoded in reasoning $r_t$.

\textbf{Probabilistic Decomposition.}
We introduce a binary latent variable $z_t \in \{0, 1\}$, where $z_t=1$ indicates satisfaction.
Although the true tool response $\mathcal{E}(o_t \mid T_t, u_t)$ is independent of the agent's internal reasoning $r_t$, we introduce $r_t$ into our world model to capture the semantic alignment between the action and the expected outcome.
By applying the law of total probability, we factorize the generation process as:
$$P_{\mathcal{W}}(o_t \mid r_t, a_t) = \sum_{z_t \in \{0, 1\}} \underbrace{P(o_t \mid a_t, z_t)}_{\text{Generation}} \cdot \underbrace{P(z_t \mid r_t, a_t)}_{\text{Intention}}.$$
Note that in the generation term, we omit $r_t$ based on the assumption that once the success status $z_t$ is determined, the specific content of $o_t$ depends primarily on the tool semantics.
This factorization leads to two specialized modules:
\begin{enumerate}
\item \textbf{Intention Predictor.} 
Estimates the probability $\rho_t$ that observation $o_t$ produced by the proposed tool call $(T_t, u_t)$ will align with the agent's intention revealed in the reasoning,  
$\tilde{\rho}_t = \mathcal{W}_\rho(z_t=1 \mid r_t, T_t, u_t).$
\item \textbf{Conditional Generator.} Generates $o_t$ conditioned on the satisfaction status,
$
\tilde{o}_t \sim \mathcal{W}_\psi(\cdot \mid [T_t, u_t], z_t).
$
\end{enumerate}

\textbf{Ideal Trajectory Simulation.}
During inference, instead of conducting stochastic sampling which risks traversing costly failure loops, we perform a deterministic simulation of the \textit{ideal trajectory}. 
Starting from the current step $t$, we construct a trajectory $\tilde{\tau}^*$ by explicitly forcing the conditional generator to satisfy the intention ($z_k=1$) at every subsequent step $k \ge t$, 
$
\tilde{o}_k \sim \mathcal{W}_\psi(\cdot \mid [T_k, u_k], z_k=1).
$

This yields a clean trajectory where every tool call works as intended and proceeds towards the solution without deviation, eliciting the \textit{latent plan} that the agent currently holds.

\textbf{Geometric Cost Calibration.}
With the latent plan $\tilde{\tau}^*$ extracted, we proceed to estimate its expected budget consumption via a \textit{pessimistic estimation} strategy. 
We observe that an agent adhering to a specific plan will persistently retry or refine arguments until the intention is met ($z_t=1$), as documented by \citet{Xue2025AnIO} and \citet{Jin2026ABF}. By modeling the number of trials as a geometric distribution with a constant initial success probability $\tilde{\rho}_k$ (ignoring potential information gain during retries), we derive an upper bound for the expected cost of each step 
$
\tilde{c}_k = \textsc{Cost}(a_k)/\tilde{\rho}_k.
$

Finally, to compare this probabilistic estimatation against the hard budget, we introduce a risk preference parameter $\gamma$. The Oracle accepts the proposed action $a_t$ if it is immediately affordable ($\textsc{Cost}(a_t) \le B_t$) and the risk-adjusted total cost remains within limits, i.e, 
$
\gamma \sum_{\tilde{a}_k \in \tilde{\tau}^*} \tilde{c}_k \le B_t.
$

Here, $\gamma$ serves as a discount factor on our pessimistic estimate, allowing the system to balance between strict safety ($\gamma \ge 1$) and aggressive goal-seeking ($\gamma < 1$).

If rejected, the feedback mechanism follows the MCO protocol but augments the returned trajectory with predicted success probabilities $\hat{o_t} = [a_t, \tilde{\rho}_t, \dots, \tilde{a}_K]$, which helps the agent identify high-risk bottlenecks for targeted re-planning.

\textbf{Simulation Reuse.}
To reduce overhead, we cache the future actions $[\tilde{a}_{t+1},\tilde{a}_{t+2},\ldots,\tilde{a}_K]$ in ideal trajectory $\tilde{\tau}^*$ upon acceptance. At the subsequent step, if the agent's proposed action $a_{t+1}$ aligns with the cached anticipation ($a_{t+1} \approx \tilde{a}_{t+1}$), we imply plan continuity. Since the remaining trajectory has already satisfied the risk-adjusted budget constraint, we bypass the simulation and grant immediate approval. We also provides other additional mechanisms to boost empirical time efficiency, see discussion in Appendix~\ref{app:mechanism}.

\section{Experiments}\label{sec:exp}
In this section, we empirically evaluate INTENT in budget-constrained and dynamic tool-market settings.
Our experiments are designed to answer three important questions:
(\romannum{1}) Can agentic models reliably satisfy hard budget constraints while solving tasks?
(\romannum{2}) How effective is INTENT compared to alternative inference-time strategies?
(\romannum{3}) How robust is INTENT to market perturbations such as price changes, new tools, and varying budgets? All experimental implementation details are provided in the Appendix~\ref{app:details}.

\subsection{Experimental Setup}

\textbf{Dataset.}
We conduct our experiments on StableToolBench \cite{Guo2024StableToolBenchTS,Guo2025StableToolBenchMirrorAPIMT}, a stable large-scale benchmark for tool learning that evolves from the widely used ToolBench \cite{Qin2023ToolLLMFL}. ToolBench provides multi-step tool-use tasks where an agent is required to iteratively generate tool calls and incorporate tool observations to complete a given instruction, and covers over 16k real-world tools collected from RapidAPI, spanning 49 diverse categories.
StableToolBench further introduces a cache-based API fallback mechanism, ensuring stable and reproducible evaluation.

Since ToolBench does not provide cost information for tools, we augment each instance with synthetic tool prices. For each query $q$, we fix a moderate budget $B = 50$, use the official retriever \cite{Reimers2019SentenceBERTSE,Qin2023ToolLLMFL} to recall 20 related tools, and then assign each tool $T^{(j)} \in \mathcal{M}$ a per-call cost $c^{(j)}$ independently sampled from a uniform distribution $U(5, 50)$, forming the market snapshot $\mathcal{M}$ for each instance $\mathcal{I} = (q, B, \mathcal{M})$. We evaluate all methods on the $765$ test instances in StableToolBench.

\textbf{Baselines.} We categorize our baselines into two groups based on whether the budget constraint is explicitly enforced during inference: \textit{Soft} baselines and \textit{Enforce} baselines.

\textit{Soft} baselines do not enforce the budget constraint, and instead rely on the model's implicit understanding of budget through prompting. This category includes: 
(\romannum{1}) \textsc{Raw}, where no cost information is provided to the model; and 
(\romannum{2}) \textsc{Prompt}, where the model is explicitly informed of tool costs and current spending via natural language prompts. These baselines evaluate the model's intrinsic ability to reason about budget without external control mechanisms.

\textit{Enforce} baselines explicitly prevent budget violations by introducing external intervention mechanisms when the agent attempts to exceed the budget. We consider three representative methods: 
(\romannum{1}) \textsc{DFSDT} \cite{Qin2023ToolLLMFL}, a heuristic depth-first search strategy that prunes branches leading to budget overflow; 
(\romannum{2}) \textsc{BTP} \cite{Zheng2024BudgetConstrainedTL}, which formulates tool selection as a multi-knapsack problem and allocates call quotas for each tool; and 
(\romannum{3}) \textsc{BATS} \cite{Liu2025BudgetAwareTE}, which employs a budget tracker to dynamically adjust agent behaviors under different remaining budget levels.

\textbf{Evaluation Metrics.}
We evaluate all methods from three perspectives: \textsc{Performance}, \textsc{Cost-Awareness}, and \textsc{Efficiency}, in order to comprehensively assess both task-solving ability and budget-sensitive behaviors.

\textsc{Performance.} These metrics evaluate the agent's ability to successfully solve tasks under budget constraints. We report:
(\romannum{1}) \textsc{Pass Rate (\textbf{PR})}, the percentage of tasks successfully solved;
(\romannum{3}) \textsc{Budget-Optimal Pass Rate (\textbf{OR})}, defined as the ratio between the number of tasks solved by the agent and the total number of tasks that are solvable under the same budget; and 
(\romannum{2}) \textsc{Win Rate (\textbf{WR})}, the proportion of tasks where the agent outperforms the reference solution.
Following the original StableToolBench, these metics are evaluated in a LLM-as-a-Judge \cite{Zheng2023JudgingLW} paradigm, where both the quality of the final answer and the tool call trace are taken into consideration.

\textsc{Cost-Awareness.} These metrics measure whether the agent exhibits awareness of budget and cost during decision making. We consider:
(\romannum{1}) \textsc{Feasible Rate (\textbf{FR})}, the proportion of tasks where the agent does not exceed the budget;
(\romannum{2}) \textsc{Average Cost (\textbf{AC})}, the average total cost incurred per task; and
(\romannum{3}) \textsc{Average Price (\textbf{AP})}, the average per-call price of selected tools, which reflects whether the agent prefers cheaper alternatives when budget is tight.

\textsc{Efficiency.} These metrics assess the computational efficiency of different methods. We adopt:
(\romannum{1}) \textsc{E2E \textbf{Time}}, the end-to-end time for completing all tasks;
(\romannum{2}) \textsc{\textbf{Lat}ency}, the average completion time per task under multi-threaded execution; and
(\romannum{3}) \textsc{\textbf{Tok}en Consumption}, the total number of tokens consumed by agents and oracles during inference. We report the relative ratio compared to the \textsc{Raw} method.

\newcommand{\deepseekicon}{\raisebox{-0.5ex}{\includegraphics[height=1.1em]{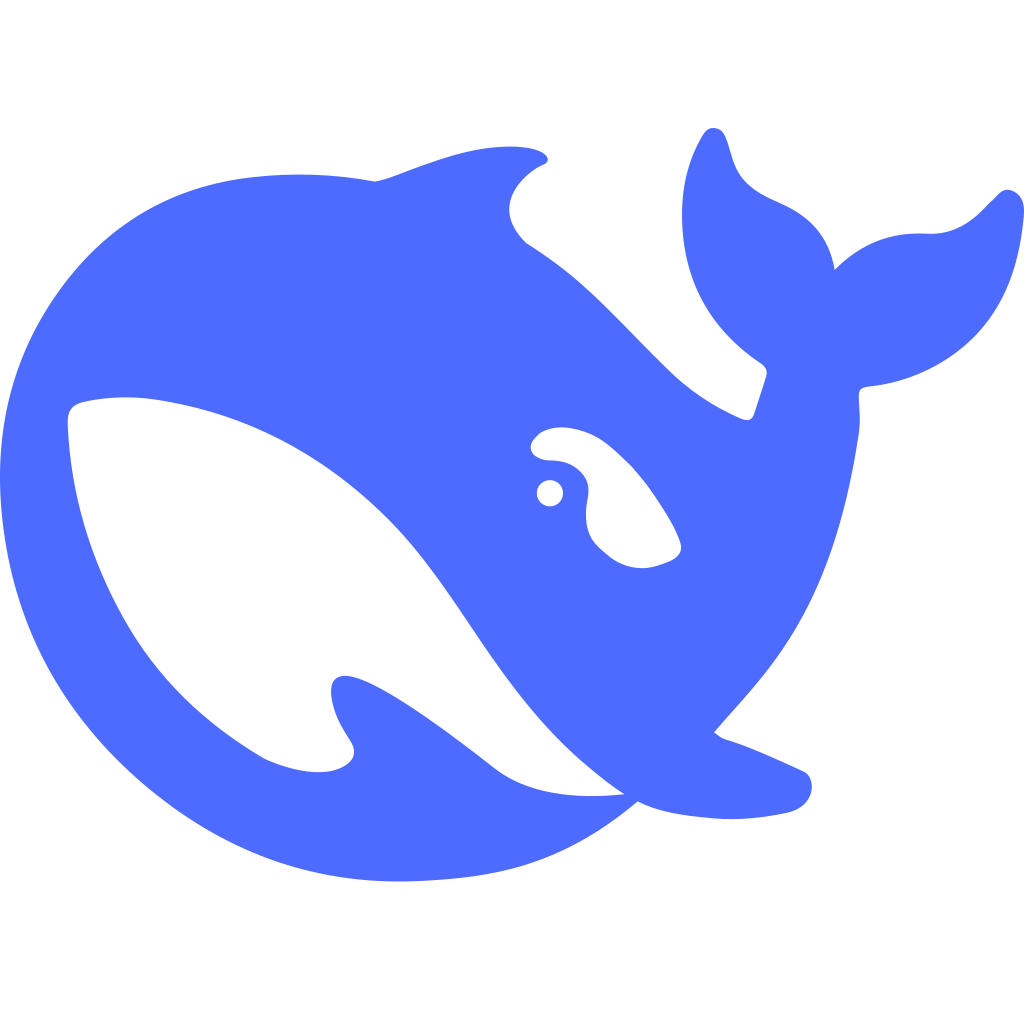}}}
\newcommand{\geminiicon}{\raisebox{-0.5ex}{\includegraphics[height=1.1em]{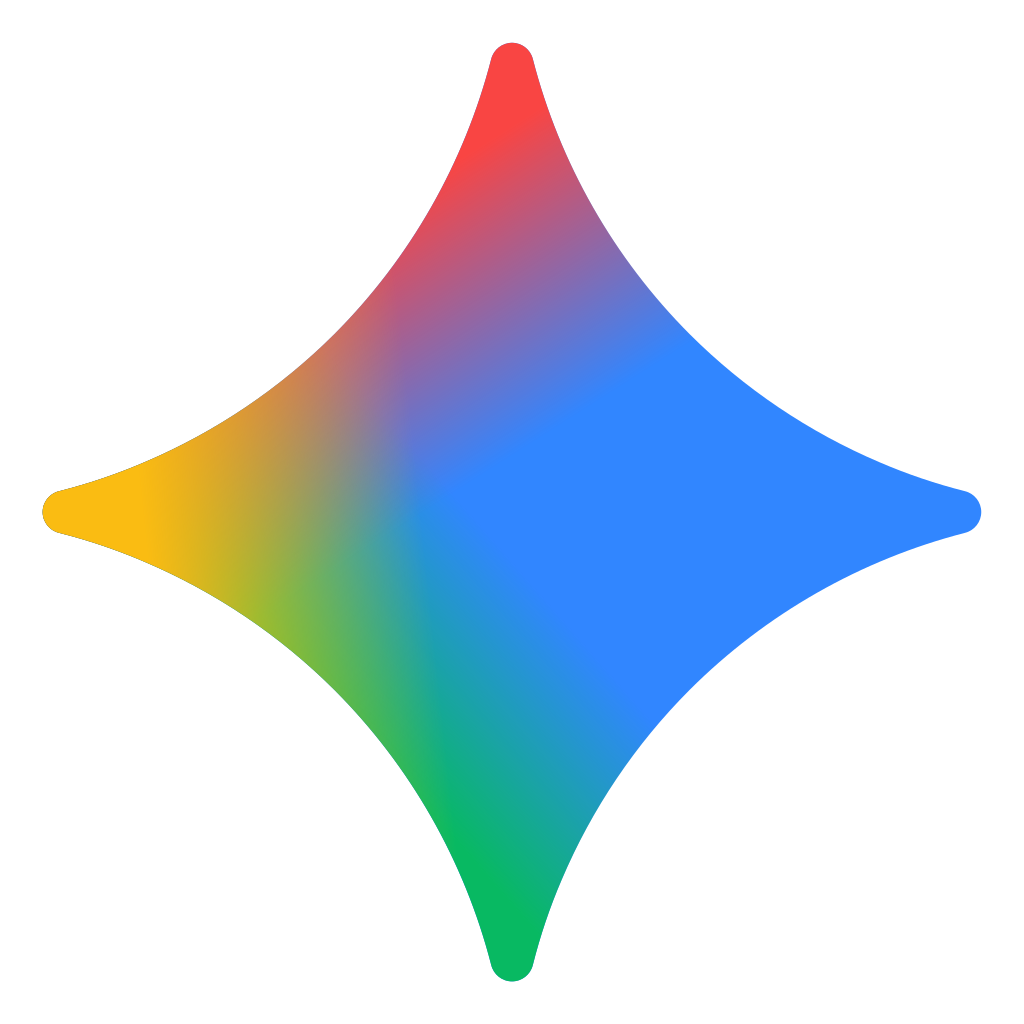}}}
\newcommand{\openaiicon}{\raisebox{-0.5ex}{\includegraphics[height=1.1em]{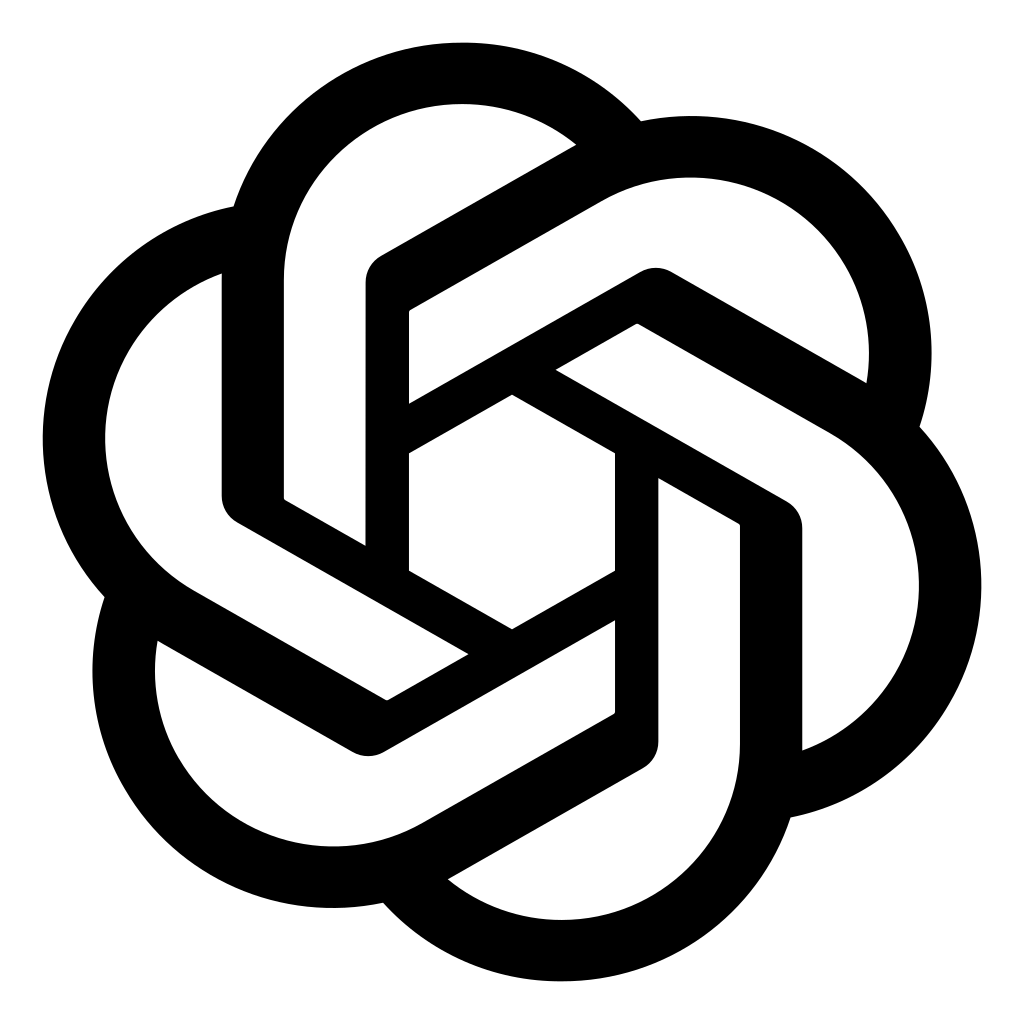}}}
\newcommand{\qwenicon}{\raisebox{-0.5ex}{\includegraphics[height=1.1em]{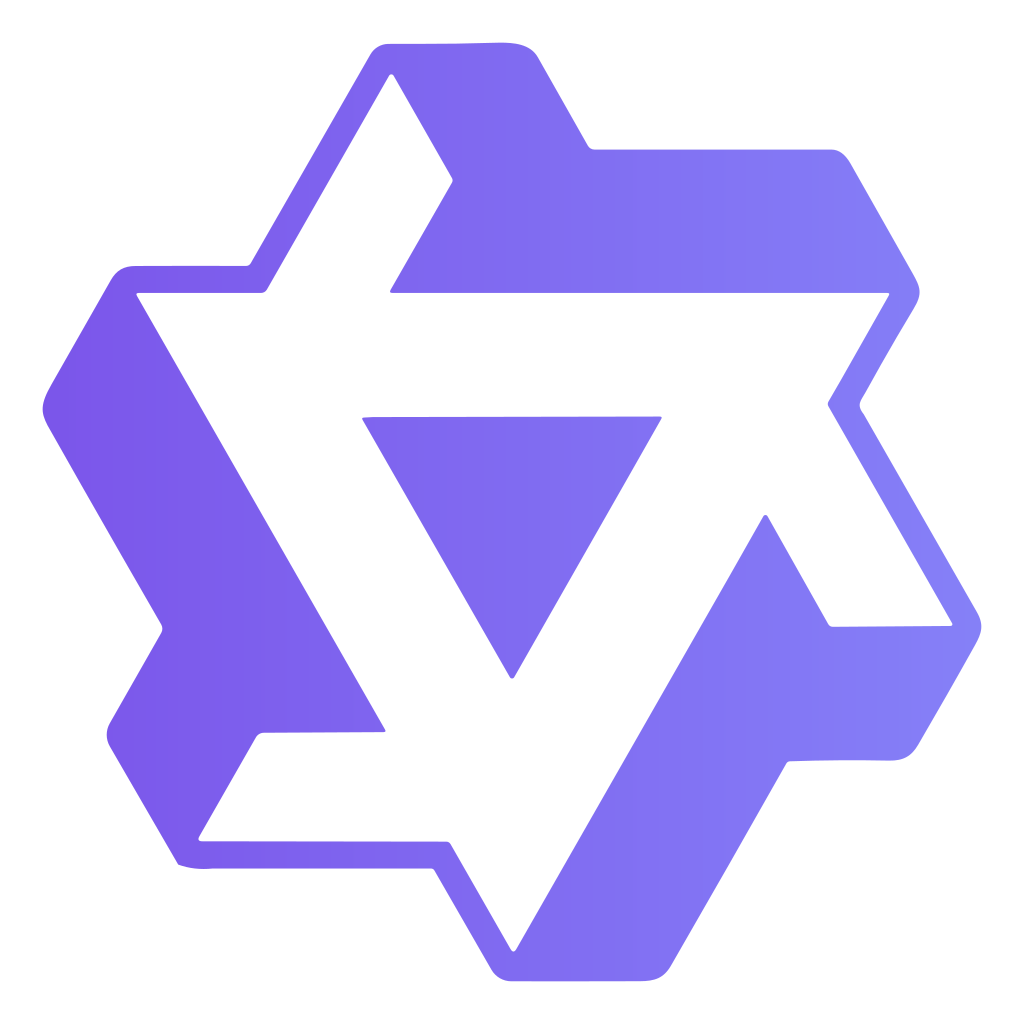}}}

\definecolor{darkgreen}{RGB}{90,121,62}
\begin{table*}[t!]
    \centering
    \caption{Main results on the cost-augmented StableToolBench \cite{Guo2024StableToolBenchTS}. We compare our proposed methods (MCO and INTENT) against \textit{Soft} (instruction-based) and \textit{Enforce}
    ({hard budget-blocking})
    baselines across both Non-Reasoning (\openaiicon GPT 4.1 mini) and Reasoning (\openaiicon GPT 5 nano) backbones. Comprehensive evaluation covers three dimensions: \textsc{Performance} (Pass Rate, Budget-Optimal Pass Rate, Win Rate), \textsc{Cost-Awareness} (Feasible Rate, Average Cost, Average Price), and \textsc{Efficiency} (E2E Time, Average Latency and Total Token Consumption relative to the \textsc{Raw} baseline). \textbf{INTENT} consistently achieves the best performance with budget adherence. {Since the benchmark partitions tasks into several predefined groups, we report the group-wise \textsc{Pass Rate} in Appendix~\ref{apx:breakdown} for completeness.}}
    \begin{adjustbox}{max width=\textwidth}
    \begin{tabular}{l l c c c c c c c c c} 
        \toprule
        \multicolumn{2}{c}{\multirow{2}{*}{\textbf{Method}}} & \multicolumn{3}{c}{\textsc{Performance}} & \multicolumn{3}{c}{\textsc{Cost-Awareness}} & \multicolumn{3}{c}{\textsc{Efficiency}} \\ 
        
        \cmidrule(lr){3-5} \cmidrule(lr){6-8} \cmidrule(lr){9-11}
        
        & & PR\,\textcolor{gray}{$\uparrow$} & OR\,\textcolor{gray}{$\uparrow$} & WR\,\textcolor{gray}{$\uparrow$} 
        & FR\,\textcolor{gray}{$\uparrow$} & AC\,\textcolor{gray}{$\downarrow$} & AP\,\textcolor{gray}{$\downarrow$} 
        & Time\,\textcolor{gray}{$\downarrow$} & Lat.\,\textcolor{gray}{$\downarrow$} & Tok.\,\textcolor{gray}{$\downarrow$} \\ 
        
        \midrule
        \multicolumn{11}{c}{\textbf{Non-Reasoning Model}} \\ 
        \midrule
        
        \multirow{2}{*}{\emph{Soft}} 
          & Raw \cite{Yao2022ReActSR}    & 19.1 & 23.3 & 37.5 & 34.5 & 102.1 & 28.2 & 1.00$\times$ & 1.00$\times$ & 1.00$\times$ \\ 
          & Prompt & 30.9 & 37.7 & 41.4 & 67.2 & 43.4 & 24.2 & 0.65$\times$ & 0.55$\times$ & 0.56$\times$ \\ 
        
        \cmidrule(lr){1-11} 
        
        \multirow{5}{*}{\emph{Enforce}} 
          & DFSDT \cite{Qin2023ToolLLMFL} & 44.9 & 54.7 & 55.4 & 100.0 & 35.2 & 20.2 & 0.97$\times$ & 0.99$\times$ & 1.14$\times$ \\ 
          & BTP \cite{Zheng2024BudgetConstrainedTL}   & 46.4 & 56.8 & 59.0 & 100.0 & 32.5 & 19.4 & 1.01$\times$ & 0.76$\times$ & 0.84$\times$ \\ 
          & BATS \cite{Liu2025BudgetAwareTE}  & 53.0 & 64.6 & 68.1 & 100.0 & 35.6 & 20.8 & 1.96$\times$ & 3.55$\times$ & 4.13$\times$ \\ 
          & MCO (Ours)    & 58.9 & 71.8 & 72.5 & 100.0 & 27.1 & 18.7 & 1.90$\times$ & 2.05$\times$ & 2.15$\times$ \\ 
        
          \rowcolor{green!15}
          \cellcolor{white} & \textbf{INTENT (Ours)} & \textbf{63.8} & \textbf{77.8} & \textbf{73.3} & \textbf{100.0} & \textbf{24.9} & \textbf{19.1} & \textbf{1.23$\times$} & \textbf{1.76$\times$} & \textbf{1.70$\times$} \\ 

        \midrule
        \multicolumn{11}{c}{\textbf{Reasoning Model}} \\ 
        \midrule
        
        \multirow{2}{*}{\emph{Soft}} 
          & Raw \cite{Yao2022ReActSR}   & 18.1 & 22.1 & 44.7 & 24.7 & 145.8 & 27.8 & 1.00$\times$ & 1.00$\times$ & 1.00$\times$ \\ 
          & Prompt & 48.5 & 59.1 & 62.0 & 87.6 & 38.1 & 20.1 & 0.64$\times$ & 0.60$\times$ & 0.76$\times$ \\ 
        
        \cmidrule(lr){1-11} 
        
        \multirow{5}{*}{\emph{Enforce}} 
          & DFSDT \cite{Qin2023ToolLLMFL} & 57.2 & 69.7 & 71.9 & 100.0 & 35.6 & 18.6 & 1.10$\times$ & 0.91$\times$ & 1.08$\times$ \\ 
          & BTP \cite{Zheng2024BudgetConstrainedTL}   & 57.7 & 70.3 & 73.2 & 100.0 & 36.2 & 18.8 & 1.22$\times$ & 1.67$\times$ & 0.87$\times$ \\ 
          & BATS \cite{Liu2025BudgetAwareTE}  & 52.8 & 64.3 & 71.7 & 100.0 & 32.6 & 17.3 & 7.67$\times$ & 10.1$\times$ & 5.76$\times$ \\ 
          & MCO (Ours)   & 71.4 & 87.0 & 82.3 & 100.0 & 31.9 & 18.1 & 1.87$\times$ & 2.23$\times$ & 2.28$\times$ \\ 
          
          \rowcolor{green!15}
          \cellcolor{white} & \textbf{INTENT (Ours)} & \textbf{76.0} & \textbf{92.6} & \textbf{86.1} & \textbf{100.0} & \textbf{29.2} & \textbf{17.9} & \textbf{1.79$\times$} & \textbf{2.16$\times$} & \textbf{2.40$\times$} \\ 

        \bottomrule
    \end{tabular}
    \end{adjustbox}
    \label{tab:main_exp}
\end{table*}

\subsection{Main Results}

Table~\ref{tab:main_exp} summarizes the main results on cost-augmented StableToolBench under both non-reasoning (\openaiicon GPT 4.1 mini) and reasoning (\openaiicon GPT 5 nano) backbones.

\textbf{Standalone agents fail to reliably respect budgets.}
Table~\ref{tab:main_exp} shows that instruction-based methods struggle under hard budget constraints. Although providing explicit cost feedback (\textsc{Prompt}) improves pass rate over \textsc{Raw}, it still violates budgets in a substantial fraction of tasks (a notable $32.8\%$ for \openaiicon GPT 4.1 mini) and remains far from the budget-optimal frontier, confirming that implicit budget awareness alone cannot prevent repetitive and unproductive tool use.

\textbf{Enforcement introduces performance-efficiency trade-offs.}
All \emph{Enforce} baselines achieve perfect feasibility as expected , yet differ markedly in effectiveness and efficiency. Heuristic pruning (\textsc{DFSDT}) and static allocation (\textsc{BTP}) are conservative, leading to limited budget-optimal pass rates. \textsc{BATS} can improve performance but incurs prohibitive inference-time overhead, particularly with reasoning models. These also highlights the limitations of classical online planning in agentic settings.

\textbf{INTENT achieves the best overall trade-off.}
Across both non-reasoning and reasoning backbones, \textbf{INTENT} consistently attains the highest pass rate while strictly respecting budgets. Notably, these gains are achieved with only moderate inference-time overhead, validating intention-level inference-time planning as an effective and practical solution for budget-aware tool use.

{
We further provide results of fine-granularity ablation of oracle components and efficiency mechanisms, as well as sensitivity analysis of risk preference parameter $\gamma$, in Appendix~\ref{apx:further_exp}. These results highlight the synergy among INTENT’s core components, while the risk preference parameter enables a effective and flexible performance–cost trade-off.}

{
Moreover, to elucidate why the prompt-based approach fails and how INTENT guides the agent toward a feasible solution path, we provide a detailed case study in Appendix~\ref{apx:case}.}

\subsection{Robustness under Dynamic Market}

To further evaluate INTENT in realistic and non-stationary environments, we design experiments to study its robustness under three forms of market dynamics: the emergence of new tools, relative price changes, and varying budget levels.

\begin{figure}[t]
\centering
\begin{subfigure}[t]{0.49\linewidth}
    \centering
    \includegraphics[width=\linewidth]{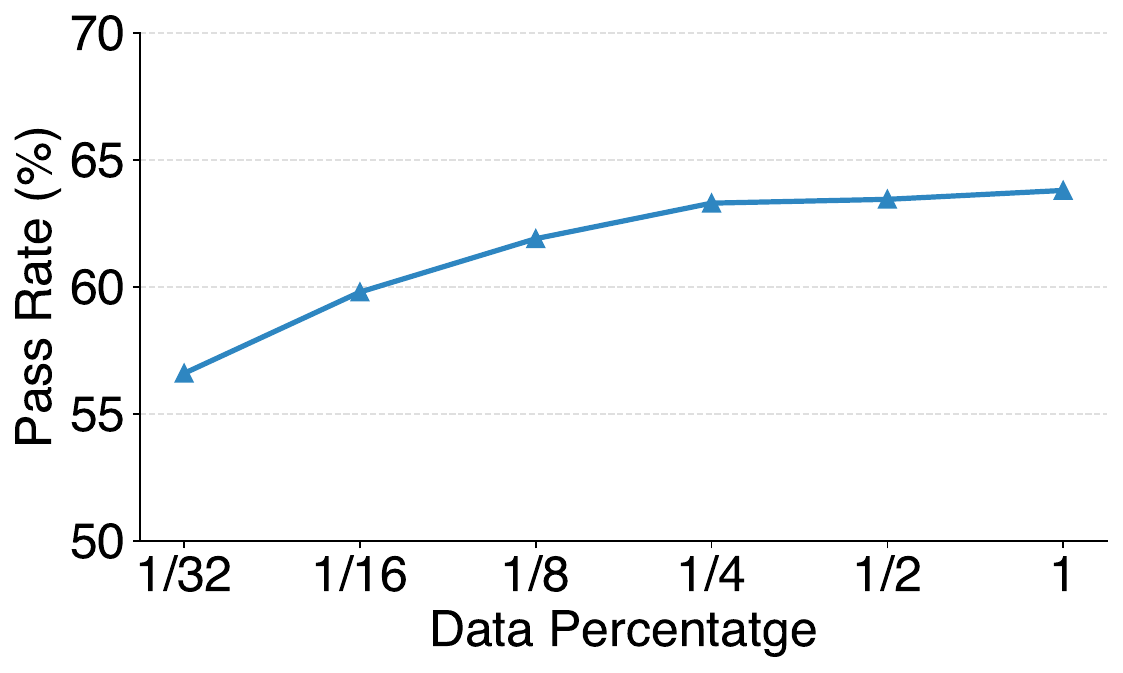}
    \caption{GPT 4.1 mini}
\end{subfigure}\hfill
\begin{subfigure}[t]{0.49\linewidth}
    \centering
    \includegraphics[width=\linewidth]{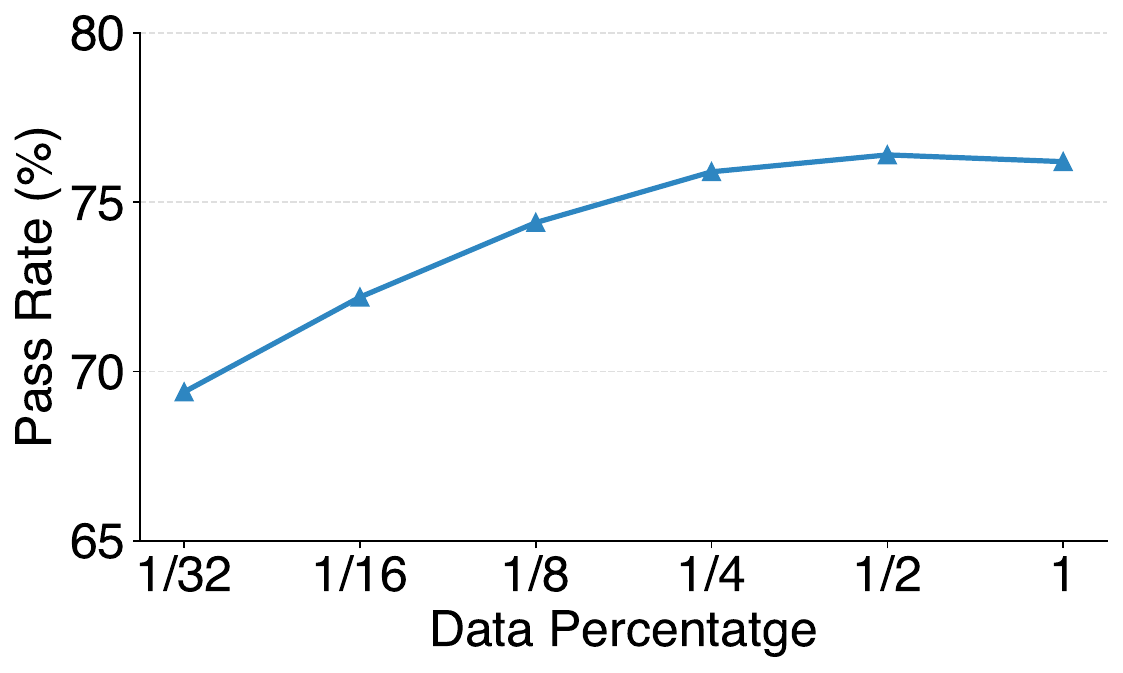}
    \caption{GPT 5 nano}
\end{subfigure}
\caption{Performance under varying amounts of oracle training data, simulating the \textbf{introduction of new tools}.
Data points correspond to fractions of the full interaction log set (from $1/32$ to $1$).
INTENT shows a clear log-linear scaling trend and strong performance even in the low-data regime, across both backbones.}
\label{fig:data_scaling}
\end{figure}
\textbf{New tools.}
A key motivation for online planning is that real-world tool markets are continually evolving, with new tools appearing that are not covered by the agent’s prior knowledge. Relying solely on parametric knowledge can therefore introduce substantial bias. To simulate this process, we vary the number of interaction logs used to train the proposed oracle, representing different stages of market exposure. Results in Figure~\ref{fig:data_scaling} show a log-linear scaling trend in the early stage as more data is accumulated. Notably, INTENT already achieves strong performance with only a few thousand logs, indicating that it can rapidly infer tool characteristics from limited interaction data.

\begin{figure}[t]
\centering
\begin{subfigure}[t]{0.49\linewidth}
    \centering
    \includegraphics[width=\linewidth]{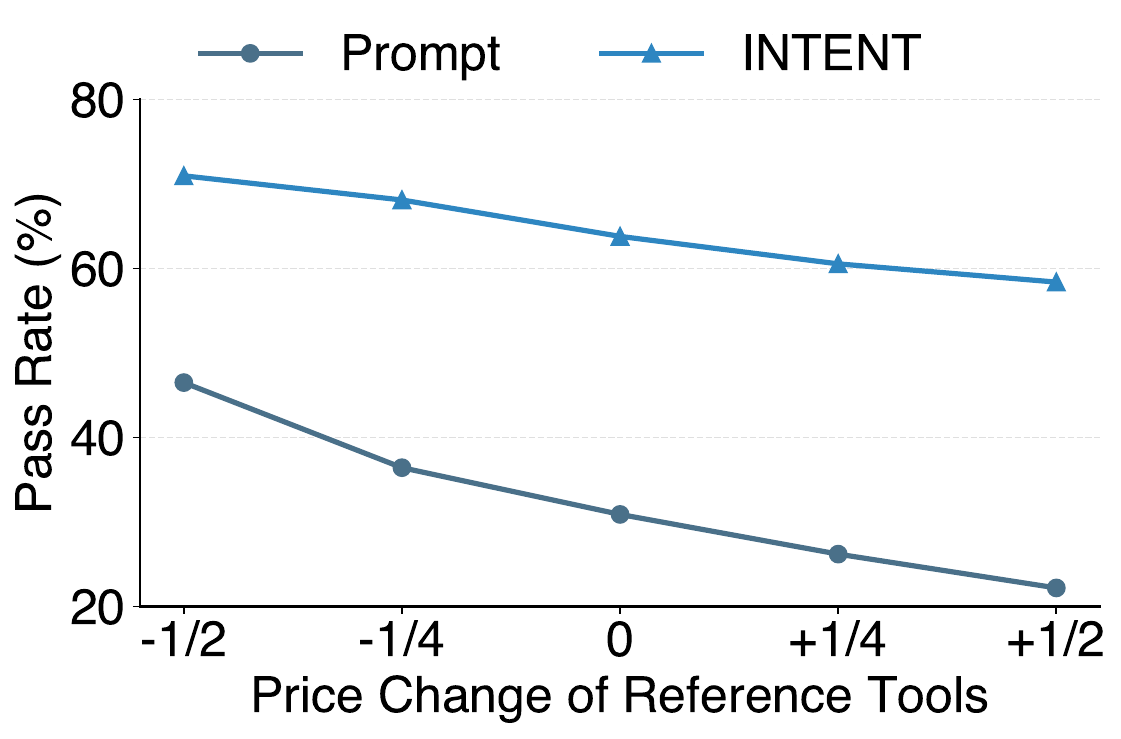}
    \caption{GPT 4.1 mini}
\end{subfigure}\hfill
\begin{subfigure}[t]{0.49\linewidth}
    \centering
    \includegraphics[width=\linewidth]{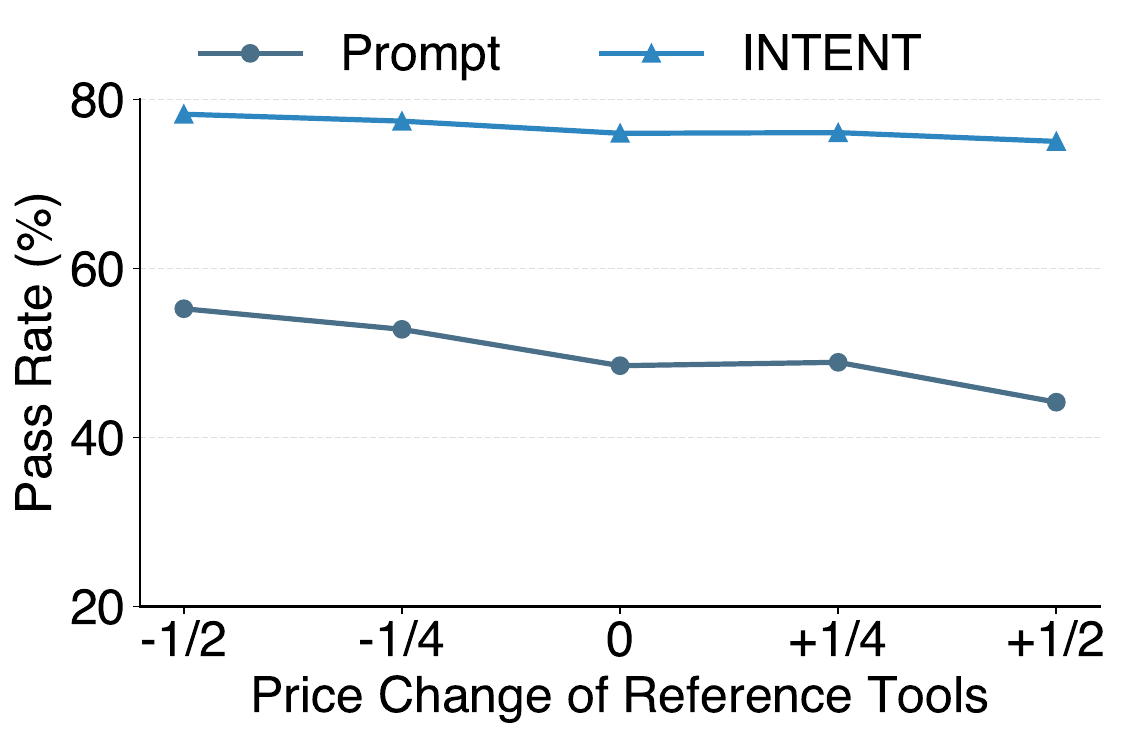}
    \caption{GPT 5 nano}
\end{subfigure}
\caption{Robustness to \textbf{relative price changes} of reference tools.
We uniformly increase or decrease the prices of reference tools by fixed ratios (from a $50\%$ discount to a $50\%$ markup), while keeping other tools unchanged.
INTENT is substantially less sensitive to price perturbations than Prompt across both backbones.}
\label{fig:price_scaling}
\end{figure}
\textbf{Price adjustments.}
In dynamic markets, tools with similar functionality often undergo relative price changes, requiring agents to reason about cost-effectiveness and substitute tools when necessary. We simulate this setting by identifying reference tools annotated in ToolBench and selectively increasing or decreasing their prices, while keeping other retrieved tools unchanged. As shown in Figure~\ref{fig:price_scaling}, \textsc{Prompt} is highly sensitive to these price perturbations, reflecting path dependence and insufficient exploration in standalone agents. In contrast, INTENT maintains consistently high pass rates with significantly smaller performance degradation, demonstrating robust cost-aware decision making.

\begin{figure}[t]
\centering
\begin{subfigure}[t]{0.49\linewidth}
    \centering
    \includegraphics[width=\linewidth]{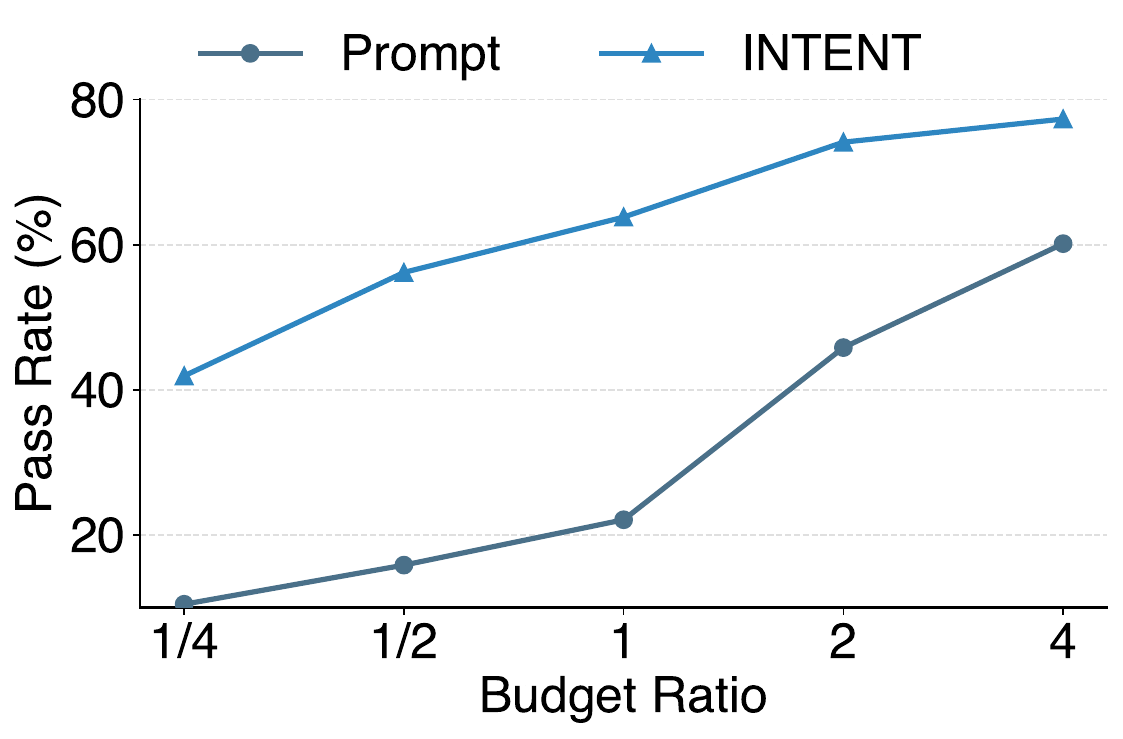}
    \caption{GPT 4.1 mini}
\end{subfigure}\hfill
\begin{subfigure}[t]{0.49\linewidth}
    \centering
    \includegraphics[width=\linewidth]{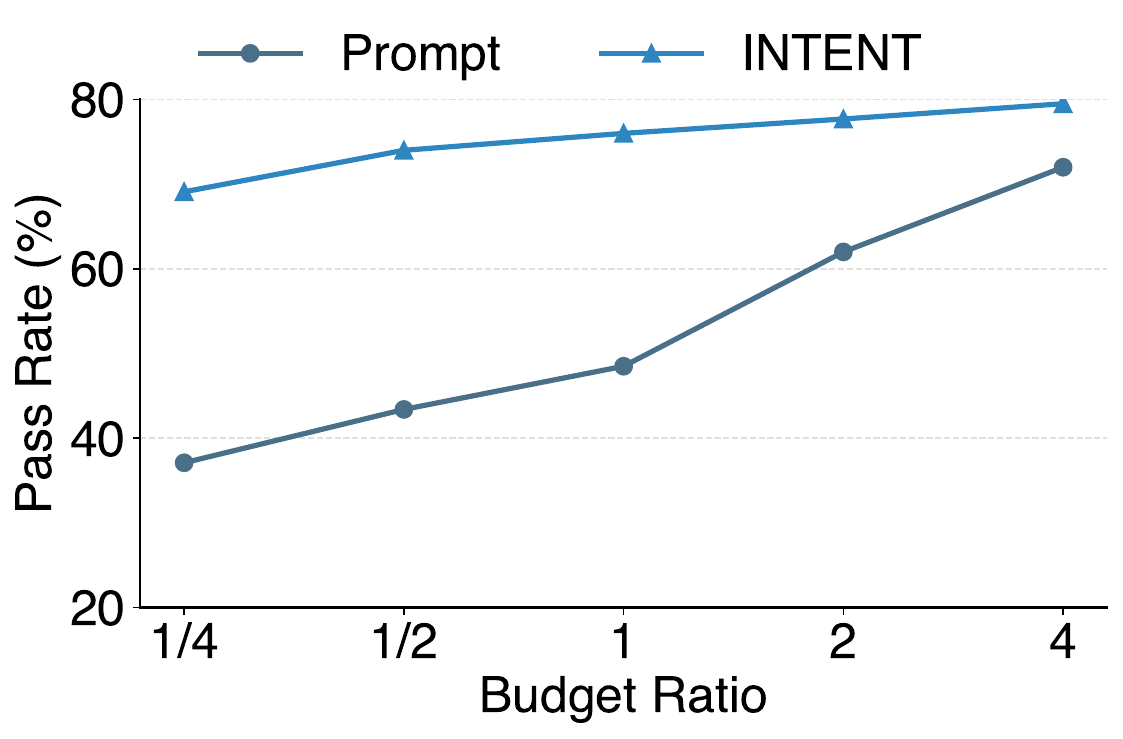}
    \caption{GPT 5 nano}
\end{subfigure}
\caption{Performance under \textbf{varying budget levels}.
Budgets are scaled by fixed ratios relative to the default setting.
INTENT scales effectively with increased budget and achieves competitive performance under tight budgets, across both backbones.}
\label{fig:budget_scaling}
\end{figure}
\textbf{Budget scaling.}
Finally, we vary the available budget to reflect different user preferences for the same task. It can be seen from Figure~\ref{fig:budget_scaling} that, although INTENT is designed for budget-constrained scenarios, it does scale when more resources are available. Moreover, its efficient budget utilization allows INTENT under tight budgets to achieve performance comparable to \textsc{Prompt} under substantially larger budgets, highlighting its ability to adapt across a wide range of resource regimes.

Overall, these results show that INTENT generalizes robustly across multiple forms of market non-stationarity, while requiring only periodic updates to a lightweight oracle model, without modifying the parameters of the underlying gigantic agentic language model.

\section{Related Work}\label{sec:related}
\textbf{Agentic AI.} Agentic AI typically refers to fronteer LLMs equipped with explicit reasoning and tool-use capabilities to autonomously solve complex, multi-step tasks. Recent systems have demonstrated strong performance in deep research and information synthesis \cite{li2025websailor,team2025tongyi}, GUI control \cite{qin2025ui}, and software engineering \cite{yang2024swe,tao2024magis}, etc.

On the infrastructure side, standardized protocols such as Model Context Protocol (MCP; \citealp{mcp_official_collection}) have enabled scalable integration of heterogeneous tools, while specialized reinforcement learning frameworks provide environments for developing and evaluating tool-augmented agents \cite{chai2025rlfactory,jiang2025verltool,fu2025areal}.

Most existing works focus on expanding the capability frontier of agents, e.g., learning to invoke a large variety of tools \cite{tang2023toolalpaca, Qin2023ToolLLMFL}, generating long-horizon tool-use trajectories \cite{chen2025reinforcement, gao2025beyond}, or coordinating multiple agents for collaborative problem solving \cite{li2025webthinker}. These methods optimize task success without constraints, and typically allow repeated tool calls until sufficient information is obtained.


In contrast, far less attention has been paid to the \emph{economic dimension} of agentic behavior.
While a growing body of work studies the efficiency of agentic systems, including token efficiency via reasoning compression or speculative decoding \cite{xia2025tokenskip,zhang2025lightthinker,chen2023accelerating,hu2025sam}, and tool efficiency through reducing or approximating tool calls \cite{xu2025alignment, nichols2025optimizing}---these approaches mainly optimize computational cost or latency.
They do not model explicit tool prices, nor do they reason about \emph{hard budget feasibility}.

As a result, existing agents cannot capture realistic scenarios in which tools are monetized, retries incur irreversible monetary costs, and agents must trade off information gain against expenditure in a dynamic tool market.
In this work, we explicitly formalize budget-constrained tool use as a sequential decision problem with hard monetary constraints, and study how a general-purpose agent can operate rationally under such conditions.





\textbf{Language World Models.} Language world models (LWMs) are trained to simulate environment dynamics in context space, enabling agents to reason about future outcomes without interacting with the real environment. In training, LWMs have been used to replace expensive or unstable external tools, significantly reducing data collection costs \cite{Guo2025StableToolBenchMirrorAPIMT, sun2025zerosearch}. Examples include simulated search engines \cite{fan2025ssrl, zhang2026searchgym}, synthesized compiler feedback \cite{pan2024training,cheng2026llm}, and general tool environments \cite{ren2025gtm,fang2025towards,xi2025agentgym}.

At inference time, LWMs are mainly applied to deterministic environments such as text-based games (e.g., ALFWorld, \citealp{shridhar2020alfworld}; GridWorld, \citealp{sasso2025foundation}), where they support planning via classical algorithms such as MCTS \cite{dainese2024generating,hao2023reasoning}.

Our setting is substantially more challenging: tools exhibit high stochasticity, and repeated failures often trigger costly retry loops. Instead of predicting exact future states, we introduce an intention-based world model that abstracts tool outcomes at the semantic level, focusing on whether a tool call satisfies the agent’s intention. This enables reliable cost estimation and budget-aware planning in highly uncertain environments.




\section{Conclusions}\label{sec:conclusion}
In this work, we highlighted budget-aware tool use as a fundamental yet underexplored problem in agentic AI model design. As agents increasingly rely on external tools in open and dynamic markets to make real-world impact, their ability to reason under hard resource constraints becomes essential for reliable and deployable systems.

To this end, we proposed INTENT, an intention-based inference-time planning framework that views budget control as a problem of anticipating whether future tool interactions will satisfy the agent’s high-level intent, rather than predicting exact tool outcomes. This abstraction allows effective budget enforcement with minimal intervention, without retraining or heavy search.

More importantly, our study suggests that budget awareness should be treated as a first-class objective in agentic AI, and that intention-level reasoning provides a natural interface between stochastic environments and resource-constrained decision making. We hope this work draws attention to budget-constrained agentic planning, and encourages further exploration of lightweight, inference-time control mechanisms for real-world agentic systems.

\clearpage

\section*{Impact Statement}
This paper presents work whose goal is to advance the budget-awareness of tool-using agents. There might be some potential societal consequences of our work, none
which we feel must be specifically highlighted here.



\nocite{langley00}

\bibliography{example_paper,baseline}
\bibliographystyle{icml2026}

\clearpage
\appendix
\section{Oracle Implementations}\label{app:algorithm}
\subsection{Training}

\textbf{Trajectory Collection.}
All oracle models are trained on agent interaction logs naturally collected from user conversations.
Each log corresponds to a task instance $\mathcal{I}=(q,B,\mathcal{M})$ sampled from the task distribution $\mathcal{D}$, and contains a ReAct-style trajectory
\(
\tau = [h_0, r_1, a_1, o_1, r_2, a_2, o_2, \dots, a_K],
\)
where $r_t$ is the reasoning trace, $a_t$ is the agent action, and $o_t$ is the environment observation.
No additional data collection or synthetic trajectories are introduced.

\textbf{Language World Model.}
$\mathcal{W}_\phi$ is instantiated as a large language model fine-tuned on tool interaction records.
Each training example consists of a tool call and its observed response $(a_t, o_t)$, and the model is trained by standard next-token prediction to approximate
\(
\mathcal{W}_\phi(o_t \mid [T_t, u_t]).
\)

\textbf{Latent Intention Annotation.}
The intention variable $z_t$ is not directly observable.
For each triple $(r_t, a_t, o_t)$, we apply an LLM-as-a-Judge \cite{Zheng2023JudgingLW} to determine whether the observation semantically satisfies the intention of the tool call expressed in $r_t$,
yielding a binary label $z_t \in \{0,1\}$.

\textbf{Conditional Generator.} $\mathcal{W}_\psi$ is implemented as a large language model trained on positive intention samples $(a_t, o_t)$ with $z_t=1$,
learning the conditional distribution
\(
\mathcal{W}_\psi(o_t \mid [T_t,u_t], z_t=1).
\)

\textbf{Intention Predictor.}
$\mathcal{W}_\rho$ is an encoder-only Transformer with a classification head.
It is trained as a binary classifier on $(r_t, a_t, z_t)$ using standard cross-entropy loss to estimate the success probability
\(
\tilde{\rho}_t = P(z_t=1 \mid [r_t,T_t,u_t]).
\)
Moreover, to obtain well-calibrated probabilities, we apply post-hoc temperature scaling \cite{Guo2017OnCO} on a held-out validation set.
The calibrated score $\tilde{\rho}_t$ is used for geometric cost estimation in the oracle.

\subsection{Additional Mechanisms}\label{app:mechanism}

In addition to the core oracle logic, we adopt several lightweight mechanisms in implementation to reduce redundant rollouts and unnecessary exploration. These mechanisms do not alter the underlying decision rules of the oracle, but improve computational efficiency in practical deployments. We incorporate these mechanisms in the main algorithm. 

\textbf{Rollout Cache.}
As described in the Section~\ref{subsec:intent}, we cache the future action sequence 
\(
\mathcal{C}_t = [\tilde{a}_{t+1}, \tilde{a}_{t+2}, \dots, \tilde{a}_K]
\)
obtained from the ideal trajectory $\tilde{\tau}^*$ upon acceptance.
If the agent's subsequent proposal satisfies $a_{t+1} \approx \tilde{a}_{t+1}$ (e.g., tolerant for some argument mismatches), we imply plan continuity and skip simulation, directly approving the action (as long as the action is budget feasible, i.e., $\textsc{Cost}(a_{t+1}) \leq B_t$), and then the cache queue pops to $\mathcal{C}_{t+1} \leftarrow [\tilde{a}_{t+2}, \dots, \tilde{a}_K]$, accordingly. If a cache miss happens instead, the rollout cache is emptied $\mathcal{C}_{t+1} \leftarrow []$, and will be replace by the new rollout $\mathcal{C}_{t+1} \leftarrow [\tilde{a}'_{t+2}, \dots, \tilde{a}'_K]$ if the current action $a_{t+1}$ gets approved. 

\textbf{Last Call Cache.}
Let $\bar{a}_t$ denote the last rejected action recorded at step $t$ (or $\varnothing$ if none).
Upon a rejection at step $t$, we set $\bar{a}_{t+1}\leftarrow a_t$.
If at step $t+1$ the agent proposes the same action again $a_{t+1}=\bar{a}_{t+1}$, this might be interpreted as a strong evidence that the action is necessary under current situation in the agent's view. Therefore we directly accept it (provided $\textsc{Cost}(a_{t+1})\le B_{t+1}$) without additional rollout, and then clear the cache $\bar{a}_{t+2}\leftarrow \varnothing$.

\textbf{Blacklist.}
During rollout, for any action $a_t = (\textsc{Call}, T_t, u_t)$ with predicted success probability $\tilde{\rho}_t < \delta$, we add the corresponding tool to a blacklist set
\(
\mathcal{B} = \mathcal{B} \cup \{T_t\}.
\)
All tools in $\mathcal{B}$ are permanently excluded from the candidate action space for the current task instance $\mathcal{I}$, i.e.,
\(
\mathcal{M} \leftarrow \mathcal{M} \setminus \mathcal{B}.
\)

\subsection{Algorithms}
\label{app:psudocode}

In this section, we provide the detailed pseudocode for the proposed framework and the oracle implementations.

\textbf{Budgeted-Constrained Agent.} Algorithm~\ref{alg:agent} summarizes the overall execution loop of an oracle-guided, budget-constrained agent.
At each step, the agent follows a standard ReAct-style \cite{Yao2022ReActSR} interaction pattern and proposes an action, which is then intercepted by an oracle for budget feasibility checking.
The oracle operates purely at inference time and does not modify the agent policy, but only decides whether to $\textsc{Accept}$ the costly tool call to be executed in the real environment or return a synthetic feedback signal for re-planning.
Algorithm~\ref{alg:mco} and Algorithm~\ref{alg:intent} instantiate the two concrete oracle implementations introduced in the Section~\ref{sec:method}, namely the Monte Carlo Oracle (MCO) and the proposed Intention-Based Oracle (INTENT).

Notably, in Algorithm~\ref{alg:intent}, we explicitly distinguish between the core components and the engineering optimizations. The core logic is highlighted in {\color{blue}\textbf{blue}}. The auxiliary mechanisms introduced to improve efficiency, i,e. the \textit{Rollout Cache}, \textit{Last Call Cache}, and \textit{Blacklist}, are marked in \textcolor{gray}{\textbf{gray}}.

\begin{algorithm}[t]
\caption{Oracle-Guided Budget-Constrained Agent}
\label{alg:agent}
\begin{algorithmic}[1]
\Require Task instance $\mathcal{I}=(q, B, \mathcal{M})$, Agent policy $\pi_\theta$, Environment $\mathcal{E}$
\State \textbf{Initialize:} History $h_0 \leftarrow [\texttt{System}, q, B, \mathcal{M}]$, Step $t \leftarrow 1$, Current budget $B_{t} \leftarrow B$
\While{$B_{t} \geq 0$}
    \State \textbf{Reasoning:} $r_t \sim \pi_\theta(\cdot \mid h_t)$
    \State \textbf{Action:} $a_t \sim \pi_\theta(\cdot \mid [h_t, r_t])$
    
    \If{$a_t$ is $(\textsc{Answer},y)$}
        \State \Return $y$ \Comment{Terminate and return final answer}
    \EndIf
    
    \State \textcolor{blue}{\textsc{// Consult the specific Oracle implementation (MCO or INTENT)}}
    \State $\text{decision}, \hat{o}_{t} \leftarrow \textsc{Oracle}(h_t, r_t, a_t, B_{t})$
    
    \If{$\text{decision} = \textsc{Accept}$}
        \State \textbf{Execute:} $o_t \sim \mathcal{E}(\cdot \mid a_t)$ \Comment{Real tool execution}
        \State $B_{t+1} \leftarrow B_{t} - \textsc{Cost}(a_t)$
    \Else
        \State \textbf{Intervention:} $o_t \leftarrow \hat{o}_{t}$ \Comment{Oracle returns simulated failure trace}
        \State \textcolor{gray}{\textsc{// Agent will re-plan based on this feedback in next iter}}
    \EndIf
    
    \State \textbf{Update:} $h_{t+1} \leftarrow [h_t, r_t, a_t, o_t]$
    \State $t \leftarrow t + 1$
\EndWhile
\State \Return \textsc{Failure} \Comment{Budget exhausted}
\end{algorithmic}
\end{algorithm}

\begin{algorithm}[t]
\caption{Monte Carlo Oracle \textbf{(MCO)}}
\label{alg:mco}
\begin{algorithmic}[1]
\Function{Oracle}{$h_t, r_t, a_t, B_{t}$}
    \State \textbf{Lookahead:} Simulate trajectory $\tilde{\tau}$ starting from $h_t$
    \State $\quad \tilde{\tau} \leftarrow [a_t]$
    \While{$a \leftarrow\textsc{last}(\tilde{\tau})$ is not \textsc{Answer}}
        \State $\tilde{o} \sim \mathcal{W}_\phi(\cdot \mid a)$ \Comment{World Model Prediction}
        \State $\tilde{r}, \tilde{a} \sim \pi_\theta(\cdot \mid [h_{t}, r_t, \tilde{\tau},  \tilde{o}])$ \Comment{Agent Policy}
        \State $\tilde{\tau} \leftarrow [\tilde{\tau}, \tilde{o}, \tilde{r}, \tilde{a}]$
    \EndWhile
    
    \State \textbf{Cost Estimation:} $\textsc{Cost}(\tilde{\tau}) \leftarrow \sum_{a \in \tilde{\tau}} \textsc{Cost}(a)$
    
    \If{$\textsc{Cost}(\tilde{\tau}) \le B_{t}$}
        \State \Return $\textsc{Accept}, \varnothing$
    \Else
        \State $\hat{o}_{t} \leftarrow \textsc{ExtractActions}(\tilde{\tau})$ \Comment{Return predicted future action sequence as hints}
        \State \Return $\textsc{Reject}, \hat{o}_{t}$
    \EndIf
\EndFunction
\end{algorithmic}
\end{algorithm}

\begin{algorithm}[t]
\caption{Intention-Based Oracle \textbf{(INTENT)}}
\label{alg:intent}
\begin{algorithmic}[1]
\State \textbf{Global:} Risk Factor $\gamma$, \textcolor{gray}{Rollout Cache $\mathcal{C} \leftarrow []$, Last Rejected $\bar{a} \leftarrow \varnothing$, Blacklist $\mathcal{B} \leftarrow \emptyset$, Blacklist Threshold $\delta$, Market Snapshot $\mathcal{M}$}
\Function{Oracle}{$h_t, r_t, a_t, B_t$}
    \color{gray}
    \State\textsc{// Last Call Cache}
    \If{$a_t \approx \bar{a}$ \textbf{and} $\textsc{Cost}(a_t) \le B_t$}
        \State \textbf{Global} $\bar{a} \leftarrow \varnothing$; \Return $\textsc{Accept}, \varnothing$ \EndIf

    \State \textsc{// Rollout Cache (Simulation Reuse)}
    \If{$\mathcal{C}$ is not empty \textbf{and} $a_t \approx \mathcal{C}[0]$}
        \State Pop $\mathcal{C}[0]$
        \State \Return $\textsc{Accept}, \varnothing$ \Comment{Plan continues}
    \Else
        \State $\mathcal{C} \leftarrow []$ \Comment{Cache miss, clear cache}
    \EndIf
    \color{black}

    \State \textbf{Ideal Trajectory Simulation:}
    \State Initialize $\tilde{\tau}^* \leftarrow []$, $k \leftarrow t$, $a_k \leftarrow a_t$, $r_k \leftarrow r_t$
    \State Total Expected Cost $\sigma \leftarrow 0$
    
    \Loop
        \State \textcolor{blue}{\textsc{// 1. Intention Prediction}}
        \State $\tilde{\rho}_k \leftarrow \mathcal{W}_\rho(z=1 \mid [r_k, T_k, u_k])$

        \color{gray}
        \State \textsc{// Blacklist Update}
        \If{$\tilde{\rho}_k < \delta$} 
        \State $\mathcal{B} \leftarrow \mathcal{B} \cup \{T_k\}$ 
        \State $\mathcal{M} \leftarrow \mathcal{M}\setminus \mathcal{B} $
        \EndIf \color{black}
        
        \State \textcolor{blue}{\textsc{// 2. Geometric Cost Calibration}}
        \State $\tilde{c}_k \leftarrow \textsc{Cost}(a_k) / \tilde{\rho}_k$
        \State $\sigma \leftarrow \sigma + \tilde{c}_k$
        
        \State \textcolor{blue}{\textsc{// 3. Conditional Generation}}
        \State $\tilde{o}_k \sim \mathcal{W}_\psi(\cdot \mid [T_k, u_k], z=1)$
        \State $\tilde{\tau}^* \leftarrow [\tilde{\tau}^*, r_k, a_k, \tilde{o}_k]$
        
        \State \textbf{Agent Step:}
        \State $\tilde{r}_{k+1}, \tilde{a}_{k+1} \sim \pi_\theta(\cdot \mid [h_t,\tilde{\tau}^*],\textcolor{gray}{\mathcal{M}})$
        \If{$\tilde{a}_{k+1}$ is \textsc{Answer}}
            \State \textbf{break}
        \EndIf
        \State $k \leftarrow k+1, a_k \leftarrow \tilde{a}_{k+1}, r_k \leftarrow \tilde{r}_{k+1}$
    \EndLoop
    
    \State \textbf{Decision:}
    \If{$\textsc{Cost}(a_t) \le B_t$ \textbf{and} $\gamma \cdot \sigma \le B_t$}
        \color{gray}
        \State $\mathcal{C} \leftarrow [\tilde{a}_{t+1}, \dots, \tilde{a}_K]$ 
        \Comment{Update Rollout Cache}
        \State \color{gray}\textbf{Global} $\bar{a} \leftarrow \varnothing$\color{black}
        \State \Return $\textsc{Accept}, \varnothing$
    \Else
        \State $\hat{o}_{t} \leftarrow [a_t, \tilde{\rho}_t, \dots, \tilde{a}_K]$ \Comment{Failure risk feedback}
        \State \color{gray}\textbf{Global} $\bar{a} \leftarrow a_t$ \Comment{Record rejection}\color{black}
        \State \Return $\textsc{Reject}, \hat{o}_{t}$
    \EndIf
\EndFunction
\end{algorithmic}
\end{algorithm}

\clearpage

\section{Implementation Details}\label{app:details}

\subsection{Training and Inference Details}

\textbf{Data.}
We train all oracle components using interaction logs provided by StableToolBench \cite{Guo2024StableToolBenchTS}.
Specifically, for the \emph{Language World Model} and the \emph{Conditional Generator}, we use the MirrorAPI-Cache training split \cite{Guo2025StableToolBenchMirrorAPIMT}, which consists of approximately 100k real tool interaction trajectories collected from RapidAPI.
Following prior work, we employ GPT-4.1-mini as an LLM-as-a-judge to annotate intention satisfaction and retain 28k trajectories where the tool calls are deemed successful.

For the \emph{Intention Predictor}, we construct a separate dataset following the ToolBench Reproduction Data \cite{Qin2023ToolLLMFL}, yielding 86k $(r, a, o)$ triples.
Each triple is annotated using GPT-4.1-mini to determine whether the tool outcome satisfies the agent’s expressed intention.
All datasets are split into training, development, and test sets with an $8{:}1{:}1$ ratio.
The development split is used for checkpoint selection and post-hoc probability calibration.

\textbf{Base Models.}
Both the \textit{Language World Model} and the \textit{Conditional Generator} are instantiated using Qwen2.5-3B-Instruct \cite{Yang2024Qwen25TR}.
The \textit{Intention Predictor} is implemented based on Qwen3-0.6B-Embedding \cite{Yang2025Qwen3TR} with a additional classification head.

\textbf{Optimization.}
We employ the standard AdamW \cite{Loshchilov2017DecoupledWD} optimizer to train all components. For the \textit{Language World Model} and \textit{Conditional Generator}, we use a batch size of 64 and a learning rate of $7\times10^{-5}$, with a linear warmup over the first 5\% of steps followed by cosine decay \cite{Loshchilov2016SGDRSG}.
The \textit{Intention Predictor} is trained with batch size 32, learning rate $5\times10^{-5}$.
All models undergo BF16 \cite{Micikevicius2017MixedPT} full-parameter fine-tuning for 2 epochs.

\textbf{Infrastructure.}
All experiments are conducted on a single NVIDIA RTX Pro 6000 GPU.
\textit{World Model} and \textit{Conditional Generator} are trained with LLaMA-Factory \cite{zheng2024llamafactoryunifiedefficientfinetuning}, while the \textit{Intention Predictor} is trained and calibrated using Transformers \cite{wolf2020huggingfacestransformersstateoftheartnatural} and scikit-learn \cite{pedregosa2018scikitlearnmachinelearningpython}.
FlashAttention-2 \cite{dao2023flashattention2fasterattentionbetter} is enabled throughout.
Inference-time model serving uses vLLM \cite{kwon2023efficientmemorymanagementlarge} with a FastAPI \cite{fastapi} frontend. Inference over task sequences in the benchmark is parallelized with a thread pool of up to 40 workers.

\subsection{Algorithm Implementation}

\textbf{DFSDT \cite{Qin2023ToolLLMFL}.}
The original DFSDT implementation does not explicitly account for monetary budgets.
We augment it with a budget enforcement mechanism: if a proposed action incurs a cost exceeding the remaining budget, the corresponding branch is immediately pruned.
The search width is fixed to 10 in all experiments.

\textbf{BTP \cite{Zheng2024BudgetConstrainedTL}.}
We implement BTP on top of the DFSDT framework.
The past experience $\mathcal{E}$ is constructed from the same Reproduction Data used in our experiments.
Tool scores $\textsc{Score}(U)$ are computed following the original formulation, using GPT-4.1-mini as the evaluator.
Tool similarity $\textsc{sim}(q_u, q)$ is measured using Qwen3-0.6B-Embedding.
We enable the Blacklist mechanism with threshold $\tau = 0.15$, consistent with the original setting.

\textbf{BATS \cite{Liu2025BudgetAwareTE}.}
We faithfully reproduce the original BATS pipeline and prompts.
Since BATS assumes \textit{per-tool} budgets, we adapt it to a unified global budget constraint.
To control inference overhead, we cap the number of \textit{Self-varification Module} calls at $K=5$ per iteration.

\textbf{MCO.}
For Monte Carlo Oracle, the \textit{Language World Model} sampling temperature is set to 1.0.
We enable the \textit{Rollout Cache} to reuse simulated trajectories across steps.

\textbf{INTENT.}
For INTENT, the \textit{Conditional Generator} sampling temperature is set to 0.3.
We enable the \textit{Rollout Cache}, the \textit{Last Call Cache}, and the \textit{Blacklist} mechanism.
Across all experiments, we fix the risk preference parameter $\gamma = 0.5$ and the rejection tolerance $\delta = 0.1$.
No task-specific hyperparameter tuning is performed.

\subsection{Evaluation Details}

\textbf{Automatic Evaluation.}
\textsc{Pass Rate} and \textsc{Win Rate} are evaluated using the same logic as SoPR and SoWR in StableToolBench, with two modifications:
(\romannum{1}) Solutions are additionally required to satisfy the budget constraint, and
(\romannum{2}) The evaluator model is upgraded to GPT-4.1-mini.
Prior work has shown strong agreement between this evaluation protocol and human judgments \cite{Qin2023ToolLLMFL}.
For \textsc{Win Rate}, the reference solution is generated by the \textsc{Prompt} baseline using GPT-5-mini.
All evaluations are conducted using major voting aggregation (3 times).

\textbf{Budget-Optimal Pass Rate.}
To estimate the achievable upper bound used in the \textsc{Budget-Optimal Pass Rate} metric, we perform an exhaustive search using DFSDT with width $w=10$, powered by GPT-5-nano.
For each query, we enumerate tool-use trajectories until collecting five unique solutions that satisfy the budget constraint.
If at least one solution is judged correct by the evaluator, the query is marked as solvable under the given budget.
This procedure yields an empirical upper bound on achievable performance.

\section{Group-wise Breakdown of Pass Rate}\label{apx:breakdown}
\begin{table}[t!]
    \centering
    \caption{Group-wise Pass Rate breakdown on cost-augmented StableToolBench. The table reports disaggregated Pass Rate across 6 benchmark-defined task groups, under non-reasoning (\openaiicon GPT 4.1 mini) and reasoning (\openaiicon GPT 5 nano) backbones. }
    \begin{adjustbox}{max width=\columnwidth}
    \begin{tabular}{l c c c c c c} 
        \toprule
        \multirow{2}{*}{\textbf{Method}} & \multicolumn{6}{c}{\textsc{Pass Rate} \textcolor{gray}{$\uparrow$}} \\ 
        
        \cmidrule(lr){2-7} 
        
        & G1-C & G1-I & G1-T & G2-C & G2-I & G3-I \\ 
        
        \midrule
        \multicolumn{7}{c}{\textbf{Non-Reasoning Model}} \\ 
        \midrule
        
        Raw     & 33.6 & 14.9 & 17.2 & 21.2 & 10.9 & 8.7 \\ 
        Prompt  & 42.6 & 30.9 & 37.6 & 27.3 & 20.1 & 10.7 \\ 
        \cmidrule(lr){1-7}
        DFSDT   & 59.7 & 42.9 & 47.7 & 44.6 & 36.3 & 33.6 \\ 
        BTP     & 48.8 & 42.3 & 54.4 & 47.0 & 37.0 & 45.4 \\ 
        BATS    & 66.6 & 45.7 & 55.6 & 56.6 & 49.7 & 43.7 \\ 
        MCO     & 63.5 & 55.5 & 58.9 & 64.6 & 47.9 & 64.5 \\ 
        
        \rowcolor{green!15}
        \textbf{INTENT} & \textbf{69.2} & \textbf{58.6} & \textbf{65.9} & \textbf{68.4} & \textbf{55.0} & \textbf{63.9} \\ 

        \midrule
        \multicolumn{7}{c}{\textbf{Reasoning Model}} \\ 
        \midrule
        
        Raw     & 25.2 & 15.1 & 16.8 & 15.2 & 19.2 & 16.1 \\ 
        Prompt  & 55.9 & 48.2 & 48.6 & 54.3 & 38.1 & 37.2 \\ 
        \cmidrule(lr){1-7}
        DFSDT   & 64.2 & 51.4 & 58.0 & 56.6 & 59.4 & 50.8 \\ 
        BTP     & 67.1 & 55.2 & 55.7 & 51.9 & 54.7 & 63.1 \\ 
        BATS    & 60.2 & 46.3 & 50.1 & 57.8 & 48.7 & 54.9 \\ 
        MCO     & 75.0 & 62.7 & 69.1 & 78.6 & 71.2 & 77.9 \\ 
          
        \rowcolor{green!15}
        \textbf{INTENT} & \textbf{81.9} & \textbf{69.2} & \textbf{75.7} & \textbf{78.3} & \textbf{73.1} & \textbf{79.5} \\ 

        \midrule
        \rowcolor{gray!15} 
        \textcolor{gray}{\textbf{Max.}} & \textcolor{gray}{\textbf{84.9}} & \textcolor{gray}{\textbf{78.0}} & \textcolor{gray}{\textbf{80.7}} & \textcolor{gray}{\textbf{85.3}} & \textcolor{gray}{\textbf{79.6}} & \textcolor{gray}{\textbf{86.9}} \\ 

        \bottomrule
    \end{tabular}
    \end{adjustbox}
    \label{tab:breakdown}
\end{table}
StableToolBench \cite{Guo2024StableToolBenchTS} further partitions tasks into several predefined groups that are largely orthogonal to the cost-sensitive decision setting studied in this work. Although these groupings are not directly aligned with our modeling assumptions, we report the group-wise \textsc{Pass Rate} breakdown in Table~\ref{tab:breakdown} for completeness.

Across all task groups, oracle-based methods consistently outperform other baselines. In particular, when instantiated with the \openaiicon GPT 5 nano backbone, INTENT achieves uniformly strong performance across groups and closely tracks the empirical upper bound that marked in \textcolor{gray}{\textbf{gray}}.

\section{Further Experiments}\label{apx:further_exp}
In this section, we present additional experiments to further analyze the behavior of INTENT, including component ablations (Appendix~\ref{apx:ablation_oracle}), efficiency mechanisms (Appendix~\ref{apx:ablation_mechanism}), and sensitivity analysis (Appendix~\ref{apx:sensitivity}).
All experiments are conducted using \openaiicon GPT~4.1~mini.

\subsection{Ablation of Oracle Components}\label{apx:ablation_oracle}
\begin{figure}[t]
\centering
\includegraphics[width=0.9\linewidth]{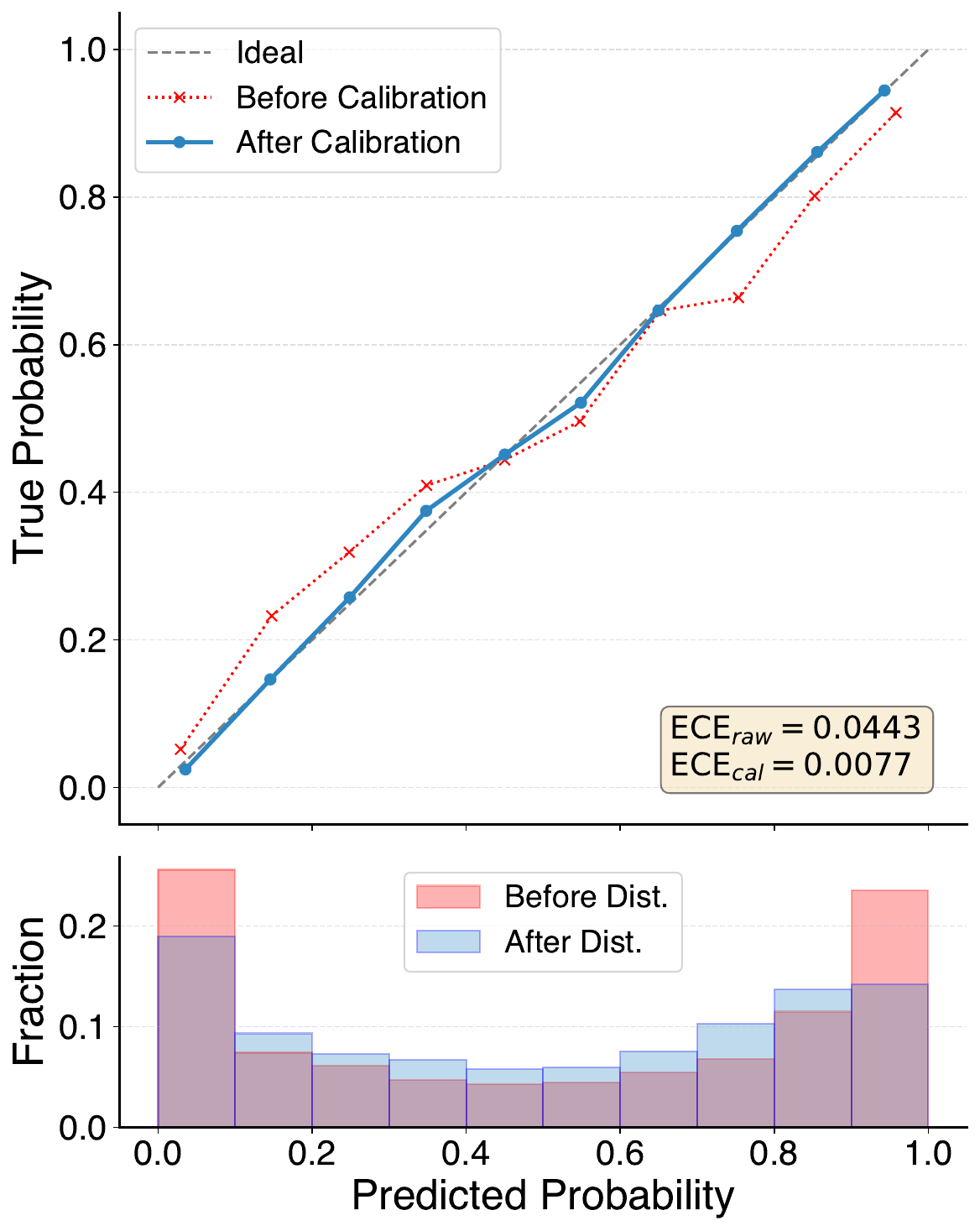}
\caption{Calibration of the intention predictor.
\textit{Top}: Reliability diagram comparing predicted probabilities with empirical success rates, before and after post-hoc calibration. Calibration substantially reduces misalignment to the ideal diagonal and improves expected calibration error (ECE).
\textit{Bottom}: Distribution of predicted probabilities before and after calibration, showing improved probability spreading without collapsing to degenerate extremes.}
\label{fig:calibration}
\end{figure}
\begin{table}[t!]
    \centering
    \caption{Component ablation of the INTENT oracle. We report the overall \textsc{Pass Rate} by ablating key components of INTENT. Removing or simplifying either modules consistently degrades performance, highlighting the complementary roles of calibrated intention estimation and ideal-trajectory generation.}
    \label{tab:ablation}
    \begin{tabular}{l c} 
        \toprule
        \textbf{Method} & PR \textcolor{gray}{$\uparrow$} \\ 
        \midrule
        
        \rowcolor{green!10}
        \textbf{INTENT} & \textbf{63.8} \\ 

        \midrule
        \multicolumn{2}{c}{\textsc{Intention Predictor}} \\ 
        \midrule
        $\hookrightarrow$ Optimistic    & 59.3 \\ 
        $\hookrightarrow$ Uncalibrated  & 59.7 \\ 
        $\hookrightarrow$ Empirical Mean       & 62.7 \\ 

        \midrule
        \multicolumn{2}{c}{\textsc{Conditional Generator}} \\ 
        \midrule
        $\hookrightarrow$ Stochastic    & 62.0 \\ 
        $\hookrightarrow$ No Rollout      & 60.2 \\ 

        $\hookrightarrow$ No Trace & 62.7 \\
        
        \bottomrule
    \end{tabular}
\end{table}
To understand the contribution of each design choice in INTENT, we conduct fine-grained ablation studies by selectively modifying the Intention Predictor and the Conditional Generator, while keeping the underlying agent and algorithmic logic unchanged. Results are summarized in Table~\ref{tab:ablation}. 

\textbf{Intention Predictor.} We consider three variants that alter how success probabilities are estimated. (\romannum{1}) \textsc{Optimistic} assumes all future tool calls succeed with probability one, i.e. $\tilde{\rho}=1$ for any tool call pairs $(\tilde{T},\tilde{u})$; (\romannum{2}) \textsc{Uncalibrated} uses the raw output probabilities of the predictor without post-hoc temperature scaling \cite{Guo2017OnCO}, see the calibration curve in Figure~\ref{fig:calibration}; (\romannum{3}) \textsc{Empirical Mean} replaces the predictor with a non-parametric baseline that assigns each tool a fixed predicted success probability equal to its empirical success frequency in the training data, ignoring contextual reasoning $\tilde{r}$ and arguments $\tilde{u}$.

\textbf{Conditional Generator.} We evaluate three alternatives that remove or weaken the lantent plan elicitation. (\romannum{1}) \textsc{Stochastic} replaces intention-satisfying rollouts with stochastic sampling from the language world model, analogous to MCO; (\romannum{2}) \textsc{No Rollout} disables lookahead entirely and only checks whether the immediate action satisfies the risk-adjusted cost constraint, without simulating future steps. 
{
(\romannum{3}) \textsc{No Trace} removes the oracle-predicted trajectory upon rejection, leaving the agent with only budget-related signals.} 

We observe consistent performance degradation when either component is ablated, confirming that both calibrated intention prediction and ideal trajectory simulation are essential for stable and effective budget-aware planning.

\subsection{Ablation of Additional Mechanisms}\label{apx:ablation_mechanism}
\begin{table}[t!]
    \centering
    \caption{Impact of Additional Mechanisms. Removing rollout caching, last-call caching, or blacklisting leads to substantially higher end-to-end time and latency, while having limited impact on Pass Rate, indicating that these mechanisms primarily improve efficiency without altering the core decision logic.}
    \begin{adjustbox}{max width=\columnwidth}
    \begin{tabular}{l | c | c c c} 
        \toprule
        \multirow{2}{*}{\textbf{Method}} & \multirow{2}{*}{PR \textcolor{gray}{$\uparrow$}} & \multicolumn{3}{c}{\textsc{Efficiency}} \\ 
        \cmidrule(lr){3-5}
        & & Time \textcolor{gray}{$\downarrow$} & Lat. \textcolor{gray}{$\downarrow$} & Tok. \textcolor{gray}{$\downarrow$} \\ 
        \midrule
        
        \rowcolor{green!10}
        \textbf{INTENT} & \textbf{63.8} & \textbf{1.23$\times$} & \textbf{1.76$\times$} & \textbf{1.70$\times$} \\ 

        \midrule
        \multicolumn{5}{c}{\textsc{Additional Mechanisms}} \\ 
        \midrule
        
        w/o Rollout Cache    & 65.3 & 1.50$\times$ & 2.66$\times$ & 1.78$\times$ \\ 
        w/o Last Call Cache  & 63.4 & 1.46$\times$ & 2.57$\times$ & 1.69$\times$ \\ 
        w/o Blacklist        & 62.5 & 1.84$\times$ & 2.65$\times$ & 1.73$\times$ \\ 

        \bottomrule
    \end{tabular}
    \end{adjustbox}
    \label{tab:mechanism}
\end{table}
Table~\ref{tab:mechanism} reports an ablation study on the additional mechanisms used in INTENT. Removing any individual mechanism leads to a noticeable degradation in efficiency, reflected by increased end-to-end time and latency, while the overall task performance remains largely stable.

\subsection{Sensitivity Analysis}\label{apx:sensitivity}
\begin{figure}[t]
\centering
\includegraphics[width=0.8\linewidth]{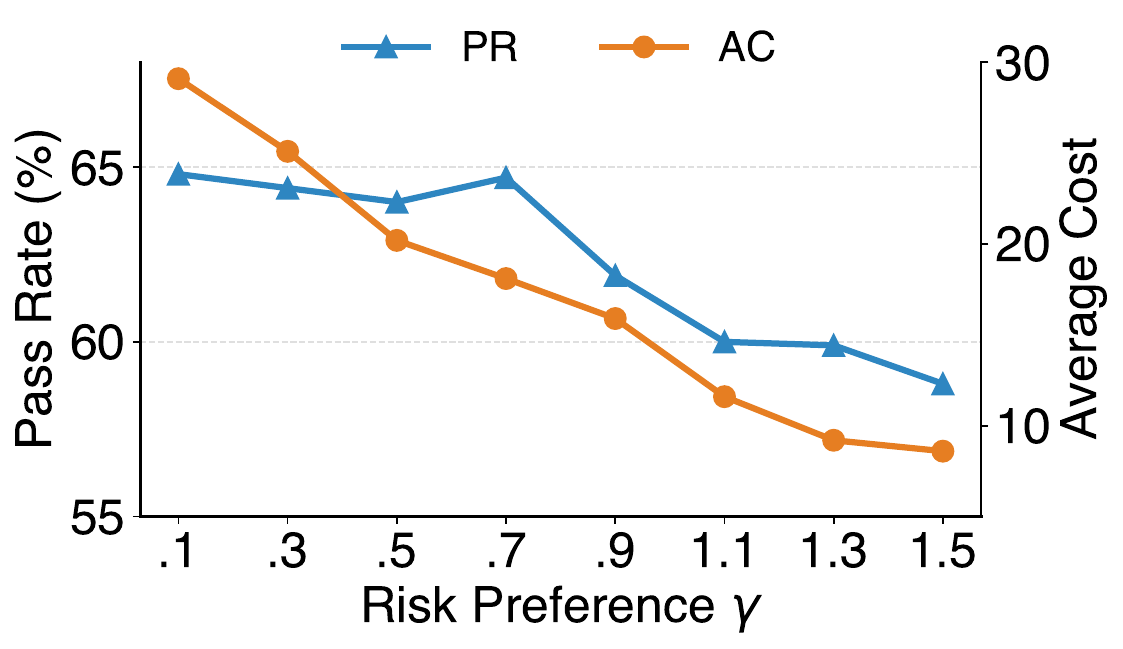}
\caption{Sensitivity to risk preference $\gamma$. Pass Rate and Average Cost as functions of the risk preference parameter $\gamma$. Increasing $\gamma$ leads to more conservative behavior with lower cost but reduced task success, revealing a clear performance–cost trade-off.}
\label{fig:risk_preference}
\end{figure}
We analyze the sensitivity of INTENT to the risk preference parameter $\gamma$, which controls the degree of conservatism in cost estimation. A larger $\gamma$ penalizes uncertain future tool usage more heavily, leading to more conservative behavior, while a smaller $\gamma$ encourages aggressive exploration.

Figure~\ref{fig:risk_preference} illustrates the trade-off between task performance and cost efficiency as $\gamma$ varies.
As $\gamma$ increases, the average cost consistently decreases, reflecting stricter budget control, while the pass rate gradually declines due to reduced exploration.
Importantly, this trade-off is smooth and well-behaved, without abrupt performance degradation.

Notably, INTENT exhibits a broad sweet spot at intermediate values of $\gamma$ (e.g., $\gamma \approx 0.5$--$0.7$), where the agent achieves high pass rate while substantially reducing average cost, indicating that INTENT is not overly sensitive to precise tuning of $\gamma$, and allows flexible control over the trade-off.

\definecolor{taskgreen}{RGB}{235, 250, 235}
\definecolor{taskframe}{RGB}{0, 120, 0}
\definecolor{thoughtred}{RGB}{250, 235, 235}
\definecolor{thoughtframe}{RGB}{180, 0, 0}
\definecolor{actionblue}{RGB}{235, 245, 255}
\definecolor{actionframe}{RGB}{0, 80, 160}
\definecolor{obsyellow}{RGB}{255, 253, 230}
\definecolor{obsframe}{RGB}{180, 150, 0}

\definecolor{simthought}{RGB}{252, 245, 245}
\definecolor{simthoughtframe}{RGB}{220, 160, 160}
\definecolor{simaction}{RGB}{245, 250, 255}
\definecolor{simactionframe}{RGB}{160, 190, 220}
\definecolor{simobs}{RGB}{255, 254, 245}
\definecolor{simobsframe}{RGB}{220, 200, 140}
\tcbset{
    oracle_style/.style={
        colback=gray!2,
        colframe=gray!40,
        width=0.96\columnwidth,
        center,
        fonttitle=\bfseries\small,
        boxrule=1pt,
        rounded corners,
        top=2mm, bottom=2mm
    }
}

\section{Case Study}\label{apx:case}

To provide a concrete and intuitive illustration of how INTENT operates, in this section, we present a detailed case study based on a representative task instance (ID: 70610) from the cost-augmented StableToolBench \cite{Guo2024StableToolBenchTS}. Agents are instantiated by \openaiicon GPT~4.1~mini.

This instance corresponds to a financial analysis scenario, where the user asks the agent to gather relevant information about a target stock. The agent is presented with 30 available tools, and multiple tool chains can potentially solve the task, leading to a combinatorially large space of feasible strategies. The full task description is provided below.
\begin{figure}[h]
    \centering
    \begin{tcolorbox}[
    colback=taskgreen, colframe=taskframe,
    title=Task Description, fonttitle=\bfseries,
    width=\columnwidth, boxrule=1pt, rounded corners
]
    \begin{tabularx}{\textwidth}{X}
    \textbf{Query:} My friend is interested in investing in stocks and wants to know more about Google. Can you fetch the quarterly cash flow, balance sheet, income statement, and ratios for Google? It would be great to have data on net income, total assets, revenue, and current ratio to assess the company's financial health. \\
    \textbf{Budget:} 50 credit units. \\
    \textbf{Market Snapshot:} 30 Tools with heterogenous per-call costs, e.g.  \url{get\_company\_income\_statement} costs 11.
    \end{tabularx}
    \end{tcolorbox}
\end{figure}
\vspace{-10pt}

We first demonstrate how the direct \textsc{Prompt} approach fails under hard budget constraints, despite having access to explicit cost information (Appendix~\ref{apx:prompt_fail}). We then show how INTENT intervenes at inference time, guiding the agent toward a feasible and ultimately successful solution through intention-aware planning (Appendix~\ref{apx:intent_sol}).

\subsection{Failure of Direct Prompt}\label{apx:prompt_fail}

In \textsc{Prompt}, the budget constraint is explicitly incorporated into the user query, and the per-call cost of each tool is appended to its description. The agent is therefore fully aware of the available budget and the monetary implications of each tool invocation.
Despite this explicit information, the agent adopts a sequential data-collection strategy, beginning with the most expensive cash flow retrieval.
\begin{tcolorbox}[
    colback=thoughtred, colframe=thoughtframe,
    title=Thought 1, fonttitle=\bfseries,
    width=\columnwidth, boxrule=1pt, rounded corners
]
\begin{tabularx}{\textwidth}{X}
I'll start by fetching Google's quarterly cash flow data. This will help in assessing the company's cash management and overall financial health. I need to be mindful of the 50 credits budget.
\end{tabularx}
\end{tcolorbox}

\begin{tcolorbox}[
    colback=actionblue, colframe=actionframe,
    title=Action 1 (\textsc{Tool Call}), fonttitle=\bfseries,
    width=\columnwidth, boxrule=1pt, rounded corners
]
\begin{tabularx}{\textwidth}{X}
\url{get\_company\_quarterly\_cash\_flow(symbol="GOOGL")} \hfill \textit{Cost: 38}
\end{tabularx}
\end{tcolorbox}

\begin{tcolorbox}[
    colback=obsyellow, colframe=obsframe,
    title=Observation 1, fonttitle=\bfseries,
    width=\columnwidth, boxrule=1pt, rounded corners
]
\begin{tabularx}{\textwidth}{X}
\url{ {"symbol": "GOOGL", "cash_flow": [{"date": "2023-01-31", "net_income": 7600000000, "operating_cash_flow": 15200000000, "capital_expenditures": 4500000000, "free_cash_flow": 10700000000},...]} } \\
\textbf{Total Spent:} 38. \textbf{Remaining Budget:} 12.
\end{tabularx}
\end{tcolorbox}

After each tool execution, the updated budget status is also injected into the observation, making the remaining budget explicitly visible to the agent.
In the subsequent reasoning step, the agent indeed acknowledges that only a limited budget remains. However, it still proceeds to invoke another expensive tool to retrieve balance sheet data. This decision ultimately pushes the total expenditure beyond the allowed budget, resulting in a failed trajectory despite the agent’s apparent awareness of the constraint.

\begin{tcolorbox}[
    colback=thoughtred, colframe=thoughtframe,
    title=Thought 2, fonttitle=\bfseries,
    width=\columnwidth, boxrule=1pt, rounded corners
]
\begin{tabularx}{\textwidth}{X}
I have obtained Google's quarterly cash flow data including net income. Next, I will use the remaining 12 credits to fetch the quarterly balance sheet to get data about total assets and current ratio for the company, as these are important to assess financial health.
\end{tabularx}
\end{tcolorbox}
\newpage

\begin{tcolorbox}[
    colback=actionblue, colframe=actionframe,
    title=Action 2 (\textsc{Tool Call}), fonttitle=\bfseries,
    width=\columnwidth, boxrule=1pt, rounded corners
]
\begin{tabularx}{\textwidth}{X}
\url{get\_company\_quarterly\_balance\_sheet(symbol="GOOGL")} \hfill \textit{Cost: 23}
\end{tabularx}
\end{tcolorbox}

\begin{tcolorbox}[
    colback=obsyellow, colframe=obsframe,
    title=Observation 2, fonttitle=\bfseries,
    width=\columnwidth, boxrule=1pt, rounded corners
]
\begin{tabularx}{\textwidth}{X}
\url{ {"Company": "Alphabet Inc.", "Symbol": "GOOGL", "Quarterly Balance Sheets": [{"Date": "2023-06-30", "Cash and Cash Equivalents": 15000000000, "Total Assets": 300000000000,  ...}, ...]} } \\
\textbf{Total Spent:} 61. \textbf{Remaining Budget:} {\color{red}\textbf{-11}}. 
\end{tabularx}
\end{tcolorbox}

\begin{tcolorbox}[
    colback=thoughtred, colframe=thoughtframe,
    title=Thought 3, fonttitle=\bfseries,
    width=\columnwidth, boxrule=1pt, rounded corners
]
\begin{tabularx}{\textwidth}{X}
Oops! The budget has been exceeded. Now I have to synthesize the final answer based on the information collected so far. 
\end{tabularx}
\end{tcolorbox}

\begin{tcolorbox}[
    colback=actionblue, colframe=actionframe,
    title=Action 3 (\textsc{Answer}), fonttitle=\bfseries,
    width=\columnwidth, boxrule=1pt, rounded corners
]
\begin{tabularx}{\textwidth}{X}
\textbf{Answer:} I fetched Google's recent quarterly financial data. Net income for the last quarters is around \$7.1B to \$7.6B. Total assets are approximately \$295B to \$300B. Cash and Cash Equivalents are roughly \$15B. The current ratio was not directly provided, but you can estimate financial health from assets and liabilities. These insights provide a view into Google's strong financial position. Note: The budget of 50 credits was exceeded by 11 credits.
\end{tabularx}
\end{tcolorbox}

\subsection{Solution Trajectory of INTENT} \label{apx:intent_sol}
\begin{tcolorbox}[colback=thoughtred, colframe=thoughtframe, title=Thought 1, fonttitle=\bfseries, width=\columnwidth, boxrule=1pt, rounded corners]
The user wants multiple financial data for Google, but the budget is 50 credits. To stay within budget, I should carefully choose which functions to call. Each function has a specific credit cost. I'll start by fetching Google's quarterly cash flow, which costs 38 credits, leaving only 12 credits left. Then, given the remaining budget, I may fetch the income statement next, which costs 11 credits. This will probably leave no credits for ratios or balance sheet. However, quarterly cash flow and income statement usually contain net income and cash flow metrics. So it's a good first step. I will fetch quarterly cash flow first for "GOOGL".
\end{tcolorbox}

\clearpage

\begin{tcolorbox}[colback=actionblue, colframe=actionframe, title=Action 1 (\textsc{Tool Call}), fonttitle=\bfseries, width=\columnwidth, boxrule=1pt, rounded corners]
\url{get_company_quarterly_cash_flow("symbol":"GOOGL")} \hfill \textit{Cost: 38, $p$:0.35}
\end{tcolorbox}

Under INTENT, the agent initially proposes a similar strategy, planning to retrieve financial data step by step. It implicitly hopes that earlier tool calls may provide sufficient coverage, potentially compensating for information that could not be obtained later due to budget limitations.

\begin{tcolorbox}[oracle_style, title=\centering \faMagic \quad \textsc{INTENT Oracle: Ideal Trajectory Simulation}]
    \footnotesize \textit{Nothing in cache, rollout begins...}

    \begin{tcolorbox}[colback=simobs, colframe=simobsframe, boxrule=0.5pt, title=Simulated Observation 1] \url{ [SomeUsefulData] }
    \end{tcolorbox}
    \begin{tcolorbox}[colback=simthought, colframe=simthoughtframe, boxrule=0.5pt, title=Simulated Thought 2]
        I have obtained quarterly cash flow data for Google including net income (10 billion), total assets (300 billion), and net assets, which are useful. However, the balance sheet is needed to get the current ratio ...
    \end{tcolorbox}
    
    \begin{tcolorbox}[colback=simaction, colframe=simactionframe, boxrule=0.5pt, title=Simulated Action 2] \url{get_company_quarterly_income_statement(symbol="GOOGL")} \hfill \textit{Cost: 11, $p$:0.97}
    \end{tcolorbox}

    \begin{tcolorbox}[colback=simobs, colframe=simobsframe, boxrule=0.5pt, title=Simulated Observation 2] \url{ [SomeUsefulData] }
    \end{tcolorbox}

    \begin{tcolorbox}[colback=simthought, colframe=simthoughtframe, boxrule=0.5pt, title=Simulated Thought 3]
        I have Google's annual income statement data showing revenue and net income for recent years, but user wanted quarterly info. The cash flow data provided quarterly info for 2023 Q2. The income statement API gave...
    \end{tcolorbox}
    
    \begin{tcolorbox}[colback=simaction, colframe=simactionframe, boxrule=0.5pt, title=Simulated Action 3]
    \textbf{Answer:} \url{[ReturnedAnswer]}
    \end{tcolorbox}
    
    \vspace{1mm}
    \centering
    \small \textbf{Risk-Adjusted Cost ($\gamma=0.5$):} $0.5 [(38/0.35) + (11/ 0.97)] = \mathbf{60.36} > 50$  \\
    \textcolor{red!80!black}{\faExclamationTriangle \ \textbf{Decision: REJECT}} (Insufficient budget for the complete latent plan).
\end{tcolorbox}

However, after performing intention-aware lookahead, the oracle determines that the projected total cost of the current plan would exceed the remaining budget. As a result, the proposed tool call is intercepted and prevented from being executed in the real environment.

\begin{figure}[h]
\begin{tcolorbox}[
    colback=obsyellow, colframe=obsframe,
    title=Observation 1 (\textsc{Oracle}), fonttitle=\bfseries,
    width=\columnwidth, boxrule=1pt, rounded corners
]
\begin{tabularx}{\textwidth}{X}
\textbf{Budget check failed.} Your current plan is predicted to exceed the remaining budget. Pay attention to the predicted success probability, some tools may have a significantly lower chance than you expected, try to avoid them in the next iteration. \url{[BudgetStateInformation]} \\
\textbf{Predicted Tool Trace:} \url{ [{"action": "get_company_quarterly_cash_flow", "price": 38, "p_success": 0.35, "expected_cost": 109.34}, {"action": "get_company_quarterly_income_statement", "price": 11, "p_success": 0.97, "expected_cost": 11.39}] } \\
Please revise your plan (fewer calls / different tools / answer directly).
\end{tabularx}
\end{tcolorbox}
\end{figure}

To facilitate more informed replanning, the oracle constructs a synthetic observation that includes the updated budget status, the anticipated sequence of future actions, and their corresponding predicted success probabilities. This feedback exposes the projected risk structure of the current plan, allowing the agent to identify potential bottlenecks and adjust its strategy accordingly.

Importantly, the trace does not reveal the predicted arguments or simulated observations of future tool calls. This design deliberately avoids injecting noisy or speculative content into the context, and prevents potential information leakage that could otherwise enable the agent to exploit the simulation as a shortcut to the final answer.

\begin{tcolorbox}[colback=thoughtred, colframe=thoughtframe, title=Thought 2, fonttitle=\bfseries, width=\columnwidth, boxrule=1pt, rounded corners]
The initial tool call is predicted to exceed the budget due to a low success probability. To reduce cost and risk, I'll switch to a cheaper tool for quarterly cash flow data. The \url{quarterly_cashflow_statements} from Freedom Finance costs only 7 credits, which is much cheaper. I can try this tool first to fetch Google's quarterly cash flow. Then, if budget allows, I will decide on the next call.
\end{tcolorbox}

\begin{tcolorbox}[colback=actionblue, colframe=actionframe, title=Action 2 (\textsc{Tool Call}), fonttitle=\bfseries, width=\columnwidth, boxrule=1pt, rounded corners]
\url{quarterly_cashflow_statements("symbol":"GOOGL")} \hfill \textit{Cost: 7, $p$:0.02}
\end{tcolorbox}

\clearpage

Based on the oracle’s feedback, the agent identifies the bottleneck in its original plan: the expensive cash flow tool has a relatively low predicted success probability in satisfying its intention, making it a high-risk and high-cost choice under the remaining budget.

Consequently, the agent revises its strategy and substitutes it with a cheaper alternative tool from a different source (Freedom Finance), trading marginal information quality for substantially improved budget feasibility.
\begin{tcolorbox}[oracle_style, title=\centering \faMagic \quad \textsc{INTENT Oracle: Ideal Trajectory Simulation}]
    \footnotesize \textit{Nothing in cache, rollout begins...}
    
    \begin{tcolorbox}[colback=simobs, colframe=simobsframe, boxrule=0.5pt, title=Simulated Observation 2] \url{ [SomeUsefulData] }
    \end{tcolorbox}
    \begin{tcolorbox}[colback=simthought, colframe=simthoughtframe, boxrule=0.5pt, title=Simulated Thought 3]
        I successfully retrieved Google's quarterly cash flow statements with the recent quarters' operating cash flow and free cash flow. The cost was only 7 credits, leaving enough credits for further data. Next, I wi...
    \end{tcolorbox}
    
    \begin{tcolorbox}[colback=simaction, colframe=simactionframe, boxrule=0.5pt, title=Simulated Action 3] \url{quarterly_income_statements(symbol="GOOGL")} \hfill \textit{Cost: 13, $p$:0.01}
    \end{tcolorbox}

    \begin{tcolorbox}[colback=simobs, colframe=simobsframe, boxrule=0.5pt, title=Simulated Observation 3] \url{ [SomeUsefulData] }
    \end{tcolorbox}

    \begin{tcolorbox}[colback=simthought, colframe=simthoughtframe, boxrule=0.5pt, title=Simulated Thought 4]
        I have successfully retrieved Google's quarterly income statement including revenue and net income for recent quarters from the freedom finance source at a cost of 13 credits. I still have about 30 credits left....
    \end{tcolorbox}
    
    \begin{tcolorbox}[colback=simaction, colframe=simactionframe, boxrule=0.5pt, title=Simulated Action 4] \url{get_company_quarterly_ratios(symbol="GOOGL")} \hfill \textit{Cost: 45, $p$:0.72}
    \end{tcolorbox}

    \begin{tcolorbox}[colback=simobs, colframe=simobsframe, boxrule=0.5pt, title=Simulated Observation 4] \url{ [SomeUsefulData] }
    \end{tcolorbox}
    \textit{...}
    
    \vspace{1mm}
    \centering
    \small \textbf{Risk-Adjusted Cost ($\gamma=0.5$):} $\mathbf{1308.34} > 50$  \\
    \textcolor{red!80!black}{\faExclamationTriangle \ \textbf{Decision: REJECT}}.
\end{tcolorbox}
\newpage

\begin{tcolorbox}[
    colback=obsyellow, colframe=obsframe,
    title=Observation 2 (\textsc{Oracle}), fonttitle=\bfseries,
    width=\columnwidth, boxrule=1pt, rounded corners
]
\begin{tabularx}{\textwidth}{X}
\textbf{Budget check failed.} \\ \url{[Prompt]} \url{[BudgetStateInformation]} \\
\textbf{Predicted Tool Trace:} \url{ [...] } \\
Please revise your plan (fewer calls / different tools / answer directly).
\end{tabularx}
\end{tcolorbox}

However, a second intention-aware lookahead reveals that the revised plan remains infeasible. Although the newly selected tool is cheaper, it exhibits low predicted intention-satisfaction probability. As a result, repeated retries are likely to be triggered, and the accumulated cost may still exhaust the remaining budget. The oracle therefore blocks this call as well, preventing the agent from entering a high-variance failure loop.

Notably, when the optional \textsc{Blacklist} mechanism is enabled, tools whose predicted success probabilities are extremely low, such as \url{quarterly_cashflow_statements} and \url{quarterly_income_statements} in this instance, are temporarily removed from the candidate tool set. This prevents the agent from repeatedly reconsidering high-risk tools in subsequent steps, further stabilizing the planning process.

\begin{tcolorbox}[colback=thoughtred, colframe=thoughtframe, title=Thought 3, fonttitle=\bfseries, width=\columnwidth, boxrule=1pt, rounded corners]
The cheaper tool for quarterly cash flow also is predicted to exceed budget due to very low success probability. Given strict budget and Oracle feedback, I should consider a tool with higher success rate or move to a single reliable call that may cover multiple requested data points. The Holistic Finance Stock Data provides annual or quarterly cash flow statements at 9 credits, which is cheaper and might have better success chance. I will try this one for quarterly cash flow.
\end{tcolorbox}

\begin{tcolorbox}[colback=actionblue, colframe=actionframe, title=Action 3 (\textsc{Tool Call}), fonttitle=\bfseries, width=\columnwidth, boxrule=1pt, rounded corners]
\url{cashflow(period":"quarterly", symbol="GOOGL")} \hfill \textit{Cost: 9, $p$:0.71}
\end{tcolorbox}

After receiving this additional feedback, the agent explores an alternative strategy and identifies a more suitable solution path. In particular, the tools provided by Holistic Finance Stock Data offer a favorable trade-off. It is relatively inexpensive and might be capable of returning aggregated financial information in a single call.

\newpage

\begin{tcolorbox}[oracle_style, title=\centering \faMagic \quad \textsc{INTENT Oracle: Ideal Trajectory Simulation}]
    \footnotesize \textit{Nothing in cache, rollout begins...}
    
    \begin{tcolorbox}[colback=simobs, colframe=simobsframe, boxrule=0.5pt, title=Simulated Observation 3] \url{ [SomeUsefulData] }
    \end{tcolorbox}
    \begin{tcolorbox}[colback=simthought, colframe=simthoughtframe, boxrule=0.5pt, title=Simulated Thought 4]
        I have fetched quarterly cash flow data for Google (GOOGL) from a cost-effective source. Next, I need to get at least one more major data piece that includes net income, total assets, revenue, and current ratio. ...
    \end{tcolorbox}
    
    \begin{tcolorbox}[colback=simaction, colframe=simactionframe, boxrule=0.5pt, title=Simulated Action 4] \url{get_company_quarterly_income_statement(symbol="GOOGL")} \hfill \textit{Cost: 11, $p$:0.94}
    \end{tcolorbox}

    \begin{tcolorbox}[colback=simobs, colframe=simobsframe, boxrule=0.5pt, title=Simulated Observation 4] \url{ [SomeUsefulData] }
    \end{tcolorbox}

    \begin{tcolorbox}[colback=simthought, colframe=simthoughtframe, boxrule=0.5pt, title=Simulated Thought 5]
        I have obtained Google's quarterly income statement with net income and revenue, but I still lack total assets and current ratio for a complete assessment. The balance sheet would provide total assets and curren...
    \end{tcolorbox}
    
    \begin{tcolorbox}[colback=simaction, colframe=simactionframe, boxrule=0.5pt, title=Simulated Action 5] \url{get_company_quarterly_balance_sheet(symbol="GOOGL")} \hfill \textit{Cost: 23, $p$:0.90}
    \end{tcolorbox}

    \begin{tcolorbox}[colback=simobs, colframe=simobsframe, boxrule=0.5pt, title=Simulated Observation 5] \url{ [SomeUsefulData] }
    \end{tcolorbox}
    \textit{...}
    
    \vspace{1mm}
    \centering
    \small \textbf{Risk-Adjusted Cost ($\gamma=0.5$):} $\mathbf{49.88} < 50$  \\
    \textcolor{green!80!black}{\faExclamationCircle \ \textbf{Decision: ACCEPT}}.
\end{tcolorbox}

Given this revised plan, after performing another rollout, the oracle determines that the revised plan now exhibits both a lower projected cost and a higher predicted intention-satisfaction probability. The corresponding risk-adjusted total expenditure falls within the remaining budget. 

Consequently, the oracle finally approves the proposed action, allowing the tool call to be executed in the real environment and the trajectory to proceed.

\newpage
\begin{tcolorbox}[colback=obsyellow, colframe=obsframe, title=Observation 3, fonttitle=\bfseries, width=\columnwidth, boxrule=1pt, rounded corners]
\url{ [{'symbol': 'GOOGL', 'date': '2023-05-05', 'period': 'Q1', 'reportedCurrency': 'USD', 'netCashFlow': 109131000000, 'freeCashFlow': 77307000000, 'cashFlowPerShare': 25.161, ...}] } \\
\textbf{Total Spent:} 9. \textbf{Remaining Budget:} 41.
\end{tcolorbox}

\begin{tcolorbox}[colback=thoughtred, colframe=thoughtframe, title=Thought 4, fonttitle=\bfseries, width=\columnwidth, boxrule=1pt, rounded corners]
I successfully fetched Google's quarterly cash flow data for Q1 2023 at a cost of 9 credits, leaving 41 credits. Next, I need to get the quarterly balance sheet (for total assets) and income statement (for revenue and net income). The income statement costs 11 credits, balance sheet 23 credits. I have enough budget to get income statement next, which is most critical for net income and revenue. Then, if budget allows, I will get the balance sheet. I will proceed with getting the quarterly income statement for Google.
\end{tcolorbox}

The actual execution outcome aligns with the agent’s expectation inferred from the ideal rollout. As a result, the agent proceeds to the next tool call following the anticipated plan.

\begin{tcolorbox}[colback=actionblue, colframe=actionframe, title=Action 4 (\textsc{Tool Call}), fonttitle=\bfseries, width=\columnwidth, boxrule=1pt, rounded corners]
\url{get_company_quarterly_income_statement(symbol="GOOGL")} \hfill \textit{Cost: 11}
\end{tcolorbox}

\begin{tcolorbox}[oracle_style, title=\centering \faMagic \quad \textsc{INTENT Oracle}]
    \textbf{Rollout Cache:} \url{[get_company_quarterly_income_statement, get_company_quarterly_balance_sheet]} \\
    \footnotesize \textit{Cache hit, allow tool call attempt.}
    \vspace{1mm}
    \centering \\
    \textcolor{green!80!black}{\faExclamationCircle \ \textbf{Decision: ACCEPT}}.
\end{tcolorbox}

As anticipated, the actual tool call matches the action recorded in the previously cached rollout (we only cache tool names in experiments for convenience). This plan continuity allows the oracle to directly approve the action without performing an additional lookahead.

\begin{tcolorbox}[colback=obsyellow, colframe=obsframe, title=Observation 4, fonttitle=\bfseries, width=\columnwidth, boxrule=1pt, rounded corners]
\url{ {"symbol": "GOOGL", "income_statement": [{"fiscal_date": "2023-09-30", "revenue": 69782000000, "operating_expense": 15500000000, ...}, ...]} } \\
\textbf{Total Spent:} 20. \textbf{Remaining Budget:} 30.
\end{tcolorbox}

\newpage

\begin{tcolorbox}[colback=thoughtred, colframe=thoughtframe, title=Thought 5, fonttitle=\bfseries, width=\columnwidth, boxrule=1pt, rounded corners]
I obtained Google's quarterly income statement showing recent revenues and net incomes at a cost of 11 credits, leaving 30 credits. The last requested data is the quarterly balance sheet for total assets and current ratio. The balance sheet call costs 23 credits, which is within the budget. I will call for the quarterly balance sheet next.
\end{tcolorbox}

\begin{tcolorbox}[colback=actionblue, colframe=actionframe, title=Action 5 (\textsc{Tool Call}), fonttitle=\bfseries, width=\columnwidth, boxrule=1pt, rounded corners]
\url{get_company_quarterly_balance_sheet(symbol="GOOGL")} \hfill \textit{Cost: 11}
\end{tcolorbox}

\begin{tcolorbox}[oracle_style, title=\centering \faMagic \quad \textsc{INTENT Oracle}]
    \textbf{Rollout Cache:} \url{[ get_company_quarterly_balance_sheet]} \\
    \footnotesize \textit{Cache hit, allow tool call attempt.}
    \vspace{1mm}
    \centering \\
    \textcolor{green!80!black}{\faExclamationCircle \ \textbf{Decision: ACCEPT}}.
\end{tcolorbox}

Similarly, the outcome of this tool call is consistent with the agent’s prior expectation under the simulated ideal trajectory. The next proposed action therefore aligns with the cached plan, resulting in another cache hit. Since the remaining sequence has already been validated to satisfy the risk-adjusted budget constraint, the oracle directly approves the action without additional rollout.

\begin{tcolorbox}[colback=obsyellow, colframe=obsframe, title=Observation 5, fonttitle=\bfseries, width=\columnwidth, boxrule=1pt, rounded corners]
\url{ [{"symbol": "GOOGL", "date": "2023-09-30", "currency": "USD", "cash_and_cash_equivalents": 154831000000, "short_term_investments": 79101000000, "net_receivables": 26484000000, ...}] } \\
\textbf{Total Spent:} 43. \textbf{Remaining Budget:} 7.
\end{tcolorbox}

\begin{tcolorbox}[colback=thoughtred, colframe=thoughtframe, title=Thought 6, fonttitle=\bfseries, width=\columnwidth, boxrule=1pt, rounded corners]
I have successfully fetched Google's quarterly balance sheet data including total assets and current assets at a cost of 23 credits. With the data from cash flow, income statement, and balance sheet, I can now calculate the current ratio and compile net income, total assets, and revenue for the user. I will now summarize all the gathered data in a final answer and finish the task.
\end{tcolorbox}

Ultimately, after gathering sufficient information, the agent terminates with a final answer, successfully completing the task within the prescribed budget.

\begin{tcolorbox}[
    colback=actionblue, colframe=actionframe,
    title=Action 6 (\textsc{Answer}), fonttitle=\bfseries,
    width=\columnwidth, boxrule=1pt, rounded corners
]
\begin{tabularx}{\textwidth}{X}
\textbf{Answer:} I have gathered the latest financial data for Google (GOOGL) to help your friend assess the company's financial health:\\\\
1. Quarterly Cash Flow (Q1 2023):\\- Operating Activities Cash Flow: \$41.68 billion\\- Investing Activities Cash Flow: \$31.12 billion\\- Financing Activities Cash Flow: -\$9.18 billion\\- Net Cash Flow: \$109.13 billion\\- Free Cash Flow: \$77.31 billion\\\\2. Quarterly Income Statement (Q2/Q3 2023):\\- Revenue (Q3 2023): \$69.78 billion\\- Net Income (Q3 2023): \$25.42 billion\\- Revenue (Q2 2023): \$76.04 billion\\- Net Income (Q2 2023): \$26.67 billion\\\\3. Quarterly Balance Sheet (Q3 2023):\\- Total Assets: \$766.94 billion\\- Total Current Assets: \$327.52 billion\\- Total Liabilities: \$608.26 billion\\- Total Shareholders' Equity: \$158.68 billion\\\\4. Financial Ratio - Current Ratio (Current Assets / Current Liabilities):\\- Total Current Liabilities can be approximated as Total Liabilities minus Shareholders' Equity: \$608.26B - \$158.68B = \$449.58 billion\\- Current Ratio = 327.52 / 449.58 $\approx$ 0.73\\\\Summary: Google has strong revenue and net income figures in recent quarters. The total assets are very large, and the current ratio of approximately 0.73 indicates that current liabilities exceed current assets, which could mean tighter short-term liquidity. However, the company also shows robust cash flows.\\\\If your friend needs more detailed financial analysis or other data, feel free to ask!
\end{tabularx}
\end{tcolorbox}

This case study illustrates that INTENT does not merely act as a budget checker that blocks infeasible actions. Instead, by exposing the risk structure of anticipated plans, it actively reshapes the agent’s search trajectory toward structurally efficient and budget-feasible solutions. Once a stable and cost-effective plan is identified, subsequent execution proceeds smoothly with minimal intervention through rollout reuse.
Together, these behaviors demonstrate how lightweight, intention-aware inference-time planning can reconcile strong agentic capability with strict monetary constraints in dynamic tool markets.

\end{document}